\documentclass[a4paper,fleqn]{cas-dc}

\usepackage[authoryear]{natbib}
\usepackage{bm}
\usepackage{xcolor}
\usepackage{colortbl}
\usepackage{fontawesome5}
\usepackage{xspace}
\usepackage{mathtools}
\usepackage{graphicx}      % 支持 \includegraphics（插入图片）
\usepackage{caption}       % 支持 \captionof（为非浮动环境添加标题）
\usepackage{booktabs}      % 支持 \toprule, \midrule, \bottomrule（表格线）
\usepackage{array}         % 支持表格列格式（如 `l`、`c` 等）
\usepackage{float}         % 支持浮动环境（如 figure* 跨栏浮动）
\usepackage{multicol}
\usepackage{soul}
\soulregister{\cite}7
\soulregister{\ref}7

\newcommand{\red}[1]{\textcolor{red}{#1}}

\newcommand{\green}[1]{\textcolor{green}{#1}}

\definecolor{Gray}{gray}{0.90}
\definecolor{mygray}{rgb}{.80,.80,.80}
\definecolor{myblue}{rgb}{0.94, 0.95, 1.0}
\definecolor{mylightblue}{rgb}{0.96, 0.97, 1.0}
\definecolor{forestgreen}{rgb}{0.0, 0.5, 0.0}
\definecolor{ashgrey}{rgb}{0.7, 0.75, 0.71}
\definecolor{darkorange}{rgb}{1.0, 0.55, 0.0}

\definecolor{backred}{RGB}{255, 190, 190}
\definecolor{backblue}{RGB}{210, 230, 250}

\newcommand{\dataset}{DVGBench}

\usepackage{tcolorbox}
\definecolor{myboxcolor}{RGB}{255,255,255} 
\definecolor{myframe}{RGB}{0,0,128} 

\newtcolorbox{mybody}{
  colback=myboxcolor,
  colframe=myframe,
  boxrule=1pt, % Adjust the border thickness
  left=1pt,
  right=1pt,
  top=1pt,
  bottom=1pt,
}

\newcommand{\icoyes}{\textcolor{forestgreen}{\faIcon{check-circle}}\xspace}

\newcommand{\icono}{\textcolor{ashgrey}{\faIcon{times-circle}}\xspace}
\begin{document}
\let\WriteBookmarks\relax
\def\floatpagepagefraction{1}
\def\textpagefraction{.001}

\shorttitle{\dataset}    

\shortauthors{Yue Zhou}  

\title [mode = title]{DVGBench: Implicit-to-Explicit Visual Grounding Benchmark in UAV Imagery with Large Vision-Language Models} 




\author[1,2]{Yue Zhou}[orcid=0000-0002-3080-6721]
%[<options>]






\author[2]{Jue Chen}%[]
\author[3]{Zilun Zhang}%[]
\author[4]{Penghui Huang}%[]
\author[4]{Ran Ding}%[]
\author[4]{Zhentao Zou}%[]
\author[5]{PengFei Gao}%[]
\author[5]{Yuchen Wei}%[]
\author[5]{Ke Li}%[]
\author[4]{Xue Yang}%[]
\author[4]{Xue Jiang}%[]
\author[1,2]{Hongxin Yang}\cormark[1]%[]
\ead{hxyang@geoai.ecnu.edu.cn}
\author[1]{Jonathan Li}%[]


\affiliation[1]{organization={Hinton STAI Institute and Key Laboratory of Geographic Information Science (Ministry of Education), East China Normal University},
            city={Shanghai},
            postcode={200241}, 
            country={China}}
            
\affiliation[2]{organization={School of Geospatial Artificial Intelligence, East China Normal University},
            city={Shanghai},
            postcode={200241}, 
            country={China}}
            
\affiliation[3]{organization={Zhejiang University},
            city={Hangzhou},
            postcode={310058}, 
            country={China}}
            
\affiliation[4]{organization={Shanghai Jiao Tong University},
            city={Shanghai},
            postcode={200240}, 
            country={China}}

\affiliation[5]{organization={Information Engineering University},
            city={Zhengzhou},
            postcode={450001}, 
            country={China}}

\cortext[1]{Corresponding author.}

\fntext[1]{This work was supported by the Shanghai Science and Technology Program (25ZR1402133, 25ZR1402268), the China Postdoctoral Science Foundation (Grant No. GZC20250239), and the National Natural Science Foundation of China (62506229).}


\begin{abstract}
Remote sensing 
(RS) large vision–language models (LVLMs) have shown strong promise across visual grounding (VG) tasks. However, existing RS VG datasets predominantly rely on explicit referring expressions—such as relative position, relative size, and color cues—thereby constraining performance on implicit VG tasks that require scenario-specific domain knowledge. This article introduces \dataset, a high-quality implicit VG benchmark for drones, covering six major application scenarios: traffic, disaster, security, sport, social activity, and productive activity. Each object provides both explicit and implicit queries. Based on the dataset, we design DroneVG-R1, an LVLM that integrates the novel Implicit-to-Explicit Chain-of-Thought (I2E-CoT) within a reinforcement learning paradigm. This enables the model to take advantage of scene-specific expertise, converting implicit references into explicit ones and thus reducing grounding difficulty. Finally, an evaluation of mainstream models on both explicit and implicit VG tasks reveals substantial limitations in their reasoning capabilities. These findings provide actionable insights for advancing the reasoning capacity of LVLMs for drone-based agents. The code and datasets will be released at \href{https://github.com/zytx121/DVGBench}{https://github.com/zytx121/DVGBench}
\end{abstract}


% \begin{highlights}
% \item 
% \item 
% \item 
% \end{highlights}

%\nocite{*}

% Keywords

\maketitle

% Main text
\section{Introduction}

\begin{figure}
    \centering
    \includegraphics[width=1.0\linewidth]{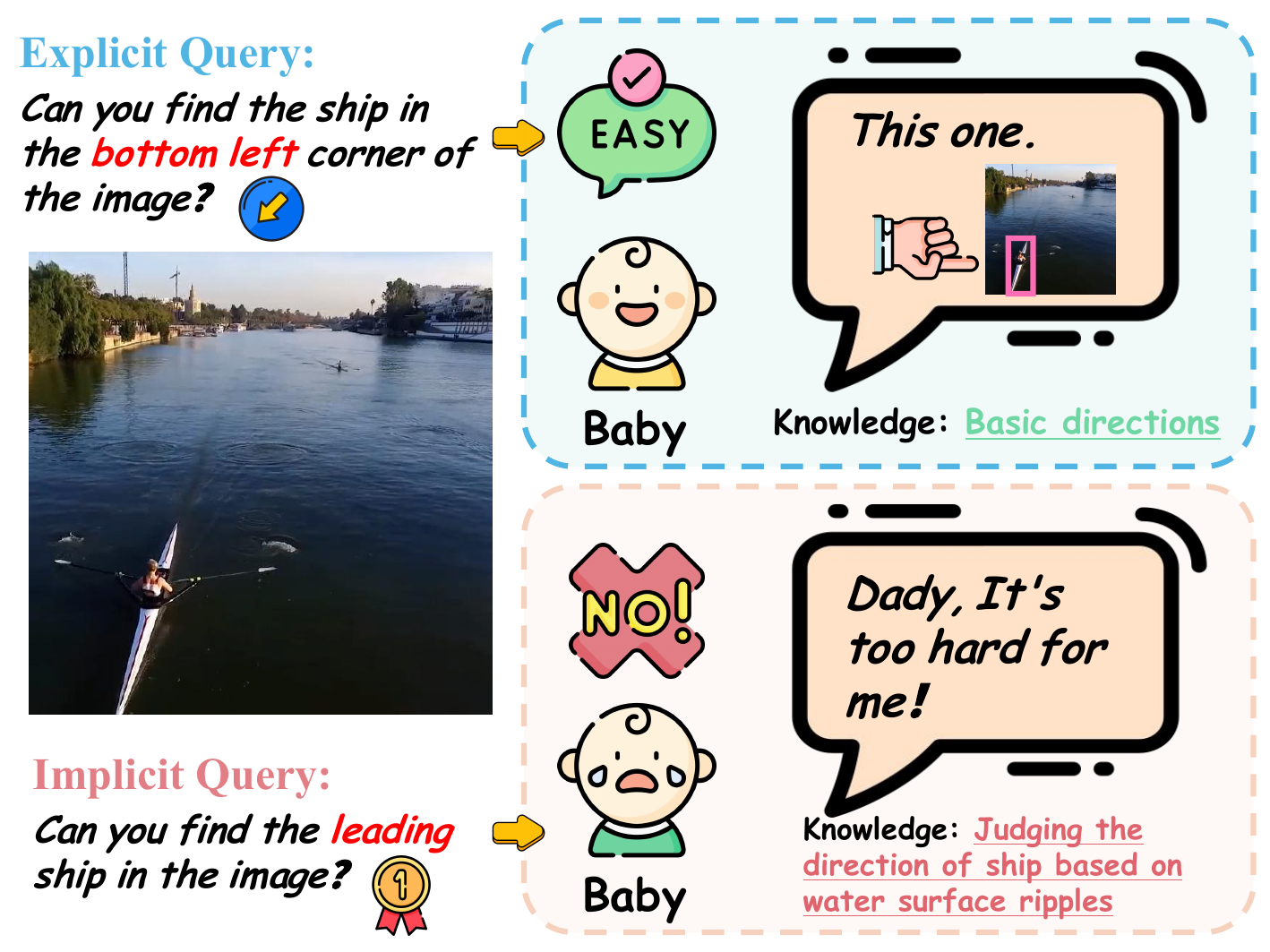}
    \caption{Infants can understand references composed of colors and relative positions easily, but cannot comprehend references involving common sense or domain knowledge. We refer to the latter as \textbf{Implicit Visual Grounding}.
    }
    \label{fig:idea}
    \vspace{-0.5cm}
\end{figure}

Visual grounding (VG) is a fundamental cognitive skill that links language to the visual world and emerges early in human development. Research in developmental psychology shows that infants can associate spoken words with visual objects during early language acquisition \citep{hollich2007young}. Cognitive science describes this capacity as a key milestone in human intelligence, forming the basis for more complex cognitive functions \citep{smallman1996foundations}. Consequently, the academic community regards the ability to handle implicit reference as a dividing line between primary and advanced intelligence \citep{osina2017out}. However, as shown in Figure~\ref{fig:idea}, we found infants can easily understand bottom left, but they struggle to judge the direction of the boat by observing the relationship between the oar and the water ripples, making it difficult for them to determine the ranking of the two boats. This type of VG task involving common sense or professional knowledge, referred to as Implicit VG, is precisely the focus of this article.

\begin{figure*}
    \centering
    \includegraphics[width=1.0\textwidth]{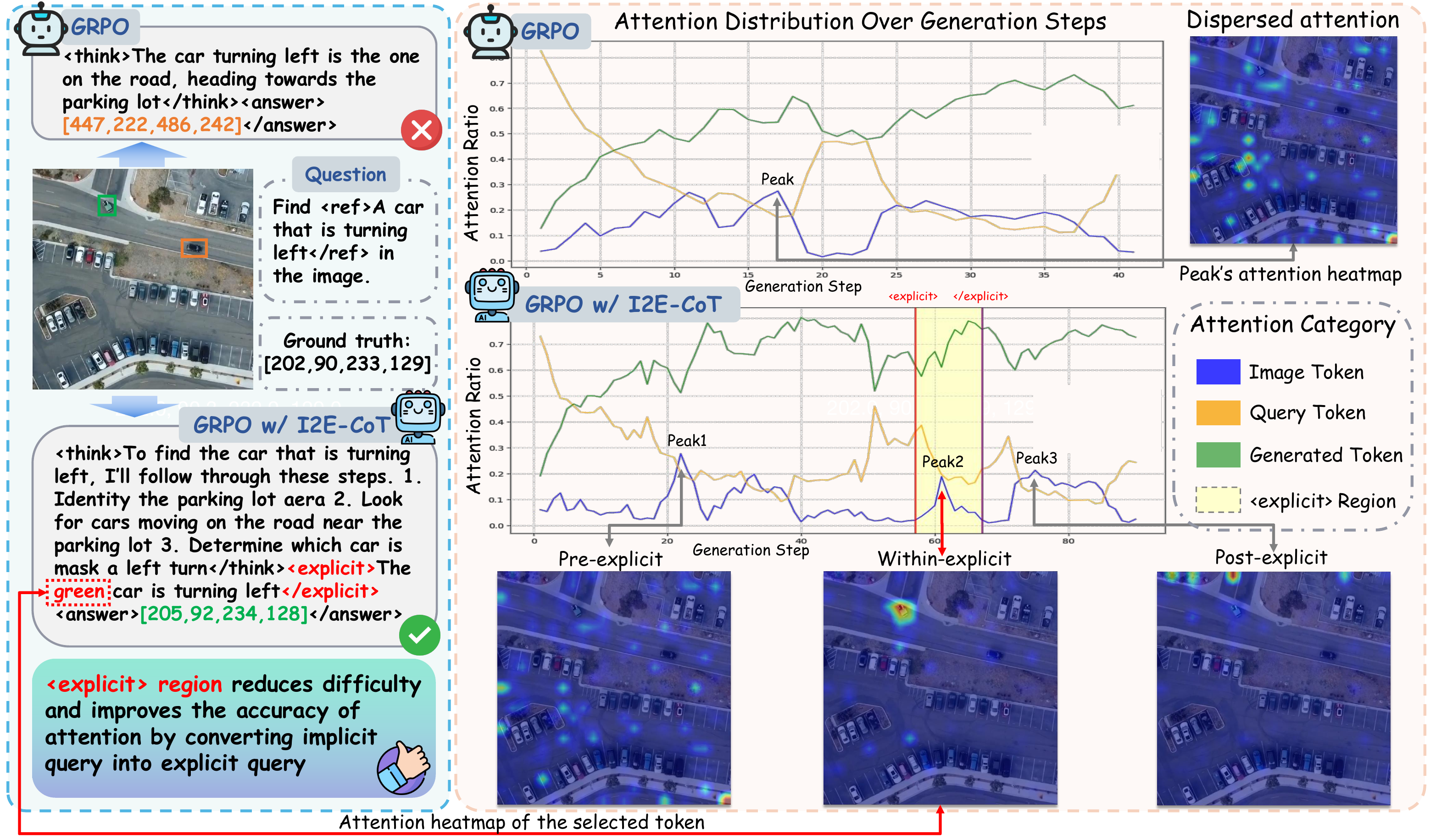}
    \caption{Overview of the Implicit-to-Explicit mechanism. This diagram compares the standard Group Relative Policy Optimization (GRPO) with our I2E-CoT approach. The GRPO mislocates the left-turning vehicle due to visual attention distraction during reasoning. In contrast, the I2E-CoT method employs the \texttt{<explicit>} token to generate an explicit reference for the object, correcting the initial localization and producing the correct answer. Attention graphs reveal that during the \texttt{<explicit>} phase, I2E-CoT identifies the explicit "green" cue, substantially increasing attention to the corresponding image tokens (blue line).
    }
    \label{fig:heatmap}
    % \vspace{-0.5cm}
\end{figure*}
% Consequently, VG is indispensable for developing low-altitude intelligent agents using drones.

The same phenomenon is also observed in Large Vision–language models (LVLMs), which have recently demonstrated strong capability on remote sensing (RS) VG tasks, spanning multiple granularities from Referring Detection (region-level)~\citep{sun2022visual,zhan2023rsvg} to Referring Segmentation (pixel-level)~\citep{yuan2024rrsis, liu2024rotated}. By learning joint representations across vision and language modalities, these models significantly enhance the accuracy of locating objects in RS imagery based on natural-language referring expressions. However, the performance superiority of LVLMs is primarily limited to tasks with explicit references, while they exhibit marked deficiencies in handling implicit VG tasks.

Existing VG datasets are largely derived from public datasets originally designed for classification or detection, lacking rich contextual information about target objects~\citep{sun2022visual,zhan2023rsvg}. This leads to annotations that emphasize explicit visual attributes—such as location and color—while struggling to include implicit descriptions requiring contextual reasoning. As a result, models trained on such data exhibit limited understanding of implicit referring expressions. Meanwhile, LVLMs also contribute to the issue: although they enable automated generation of referring data at scale, their outputs remain biased toward explicit references, as they are trained predominantly on earlier human-annotated datasets that contain mostly explicit descriptions~\citep{kuckreja2024geochat,li2024vrsbench}. This creates a "cycle of bias." Together, these factors lead to LVLMs' underperformance in implicit VG tasks.

On the other hand, the rapid proliferation of unmanned aerial vehicles (UAVs) enriches real-world application scenarios, such as traffic and disaster. VG in these scenes often involves higher-level cognition. In UAV applications, explicit references hold limited practical utility (e.g., ``the red vehicle in the top-left") — merely stating the position presupposes the answer. In contrast, implicit references that require domain knowledge are operationally meaningful (e.g., ``the vehicle making an illegal left turn against traffic" or ``people trapped by floodwater"). The significant disconnect between existing drone VG datasets and real-world application scenarios limits the practical utility of LVLMs in UAV applications.

% train a LVLM named DroneVG-R1 with strong implicit reasoning capabilities. Specifically,

To address these issues, we introduce \dataset, an implicit visual grounding benchmark built from drone imagery across 6 diverse scenes. It contains 2,863 annotated samples and features a dual-query structure for each object: one with an explicit reference (based on direct visual cues like position or color) and another with an implicit reference (requiring commonsense reasoning, e.g., detecting traffic violations). This design enables systematic evaluation of model reasoning by comparing performance on explicit vs. implicit queries. Based on \dataset, we propose an Implicit-to-Explicit Chain-of-Thought (I2E-CoT) to reduce the difficulty of implicit VG by converting implicit queries into explicit ones. Figure~\ref{fig:heatmap} illustrates the underlying mechanism of I2E-CoT. Specifically, we visualize the attention ratios between the model's output tokens, image, query, and generated tokens, and plot attention heatmaps for the peak regions of the image token ratio to analyze in detail how I2E-CoT influences the model's output \citep{yang2025look}. The attention heatmap at peak point in GRPO's attention curve show scattered focus, potentially causing erroneous outputs. In contrast, visualizations before, during, and after the explicit region reveal that it successfully concentrates attention on the object area. Thus, by generating explicit tokens, I2E-CoT enables guide the model to generate more accurate bounding box. When combined with GRPO \citep{wu2024deepseek}, I2E-CoT can further enhance the model’s reasoning ability and generalization performance. Moreover, this paper also introduces a reasoning reward based on the textual similarity of explicit references. This reward is even more critical for the VG task than the perception reward. Finally, we trained an LVLM with implicit reasoning capabilities, named DroneVG-R1, which supports both region-level and pixel-level VG tasks. Extensive experimental results validate the superiority of the proposed method for UAV implicit VG tasks. Diverse analyses provide valuable suggestions for future work.

% To address these issues, we introduce the \dataset, an implicit VG benchmark, constructed from drone images that span 6 diverse scenes. Comprising 2,863 meticulously annotated samples, \dataset distinguishes itself from conventional RS VG datasets by providing a paired query structure for each referred object: one with an explicit reference and another with an implicit reference. Explicit references denote direct visual cues such as relative spatial positions or colors within the image, whereas implicit references require reasoning with commonsense knowledge or external context, such as vehicles violating traffic regulations or soccer players potentially in an offside position. This dual-query design enables effective evaluation of model reasoning capabilities through systematic comparison of performance on explicit versus implicit references. Inspired by the observation that infants are sensitive to explicit references but not to implicit instructions, we train a model named DroneVG-R1 with strong implicit reasoning capabilities. Specifically, we propose an Implicit-to-Explicit Chain-of-Thought (I2E-CoT) technique, the core idea of which is to reduce the difficulty of VG by converting implicit queries into explicit queries. 

% To further analyze the underlying mechanism, we examine the proportion of the model's output text relative to the three types of tokens — image, query, and generated — and visualized the heatmaps corresponding to the peak image tokens \citep{yang2025look}.

\begin{table*}
% \resizebox{\columnwidth}{!}{
\caption{Comparison Between Existing RS VG Datasets and DVGBench. There are three main VG forms: horizontal bounding box (HBB), oriented bounding box (OBB), and 3D bounding box (3DBB). Text refers to the method of generating referring text. Pair: includes both explicit and implicit references.}
  \centering
  \setlength{\tabcolsep}{3pt}
  \begin{tabular}{lcccccccccc}
    \toprule
    
    Dataset & Year & Publish & Source& \# Refers & Width & Format & Text & Explicit  & Implicit & Pair  \\

    \midrule
    RSVG~\citep{sun2022visual} & 2022 & ACM MM  & satelite & 5,505 & 1,024 &  HBB &  Rule  & \icoyes & \icono  & \icono   \\
    
    DIOR-RSVG~\citep{zhan2023rsvg} & 2023 & TGRS &  satelite  & 27,133 & 800  &  HBB &  Rule &  \icoyes & \icono  & \icono   \\
    
    RSVG-HR~\citep{lan2024lqvg} & 2024 & TGRS &  satelite  & 2,650 & 1,024 &  HBB & Rule  &  \icoyes & \icono & \icono  \\
    
    OPT-RSVG~\citep{li2024lgpa} & 2024 & TGRS &  satelite  & 48,952 & 152$\sim$10,569  &  HBB & Rule  &  \icoyes & \icono & \icono  \\
    RRSIS~\citep{yuan2024rrsis} & 2024 & TGRS & satelite  & 4,420 & 512$\sim$5,616 &  Mask & Rule  &  \icoyes &\icono & \icono  \\ 
    RRSIS-D~\citep{liu2024rotated} & 2024 & CVPR & satelite  & 17,402 & 800  &  Mask & Rule  &  \icoyes &\icono & \icono \\ GeoChat~\citep{kuckreja2024geochat} & 2024 & CVPR & satelite  & 63,883 & 600$\sim$1,024  & OBB & LVLM  &  \icoyes  & \icono & \icono \\
    VRSBench~\citep{li2024vrsbench} & 2024 & NeurIPS & satelite  & 38,689 & 512 & OBB &  LVLM  & \icoyes &\icono & \icono \\ 
 
    AirSpatial-G~\citep{zhou2025airspatial} 
     & 2025 & TGRS & drone & 80,497 & 4000 & 3DBB & Rule  &  \icoyes & \icono & \icono \\

     XLRS-Bench~\citep{wang2025xlrs} 
     & 2025 & CVPR & satelite & 12,619 & $\sim$8500  &  HBB & Manual  &  \icoyes  & \icono & \icono \\
     GeoPix~\citep{ou2025geopix} 
     & 2025 & GRSM & satelite & 140,412 & 600$\sim$1,024  &  Mask &  LVLM &  \icono  & \icoyes & \icono \\
    \rowcolor{myblue}DVGBench 
    (Ours) & 2025 & - & drone & 2,863 & 540$\sim$1,920  &  Mask &  Manual &  \icoyes  & \icoyes & \icoyes  \\
    \bottomrule
  \end{tabular}
  % \vspace{0.1cm}
    % \vspace{-0.3cm}
    \label{tab:rsvg_dataset}
\end{table*}

In summary, our key contributions are as follows:

\begin{itemize}

\item \dataset, a human-annotated VG benchmark designed for real-world UAV applications, is presented. It spans six diverse scenarios and provides both box and mask-level annotations, along with explicit as well as implicit referring expressions.

\item Based on \dataset, DroneVG-R1, an LVLM tailored for implicit VG in UAV contexts, is proposed. A segmentation model is incorporated to support reasoning segmentation.

\item An I2E-CoT strategy is introduced to enhance grounding accuracy by converting implicit references into explicit textual descriptions. To incentivize this conversion, a novel reasoning reward function based on explicit reference similarity is designed.

\item Extensive evaluations of existing models are performed, uncovering their limitations in implicit VG. Through comparative analysis of performance on explicit versus implicit queries, insights into the reasoning gaps and directions for improvement are provided.
  
\end{itemize}

\section{Related Work}

VG for RS aims to localize a specified object in satellite or UAV imagery based on a natural-language description. It differs markedly from classical RS image interpretation tasks, such as object detection~\citep{yang2022detecting}, and semantic segmentation~\citep{anonymous2024textseg}, in input modality: classical tasks take only visual input, whereas VG takes both the image and a textual query. VG can be divided into two types: region-level VG \citep{qiao2020referring} and pixel-level VG \citep{lai2024lisa}. In RS, the former is termed referring detection~\citep{zhou2024geoground}, whereas the latter is termed referring segmentation~\citep{liu2024rotated}.  Table~\ref{tab:rsvg_dataset} illustrates the landscape of existing RS VG datasets. Across region-level and pixel-level annotations, DVGBench stands as the pioneering dataset that includes both explicit and implicit reference pairs.

\subsection{VG in General Domains}

Recent years have witnessed significant advancements in VG, evolving from models trained on datasets like RefCOCO ~\citep{refcoco} for single-object reference to those addressing more complex scenarios involving multiple or even no objects, as exemplified by gRefCOCO~\citep{gres}. The research scope has expanded to include robust VG~\citep{li2023irvg, wu2024rris}, which requires models to abstain from output when no referred object exists, and reasoning-aware segmentation as introduced by LISA~\citep{lai2024lisa}, demanding deeper cognitive understanding. While traditional approaches often treat VG as a detection problem~\citep{deng2023transvg, shi2024mdetr}, recent progress in LVLMs offers more flexible solutions, such as quantizing spatial coordinates into discrete tokens~\citep{peng2023kosmos, chen2023shikra} or directly predicting bounding boxes~\citep{bai2025qwen2,chen2024internvl}, with frameworks like Text4Seg~\citep{anonymous2024textseg} further enabling pixel-level grounding without structural modifications. By comparison, VG in the RS domain is still relatively underdeveloped.

\begin{figure*}
    \centering
    \includegraphics[width=1.0\textwidth]{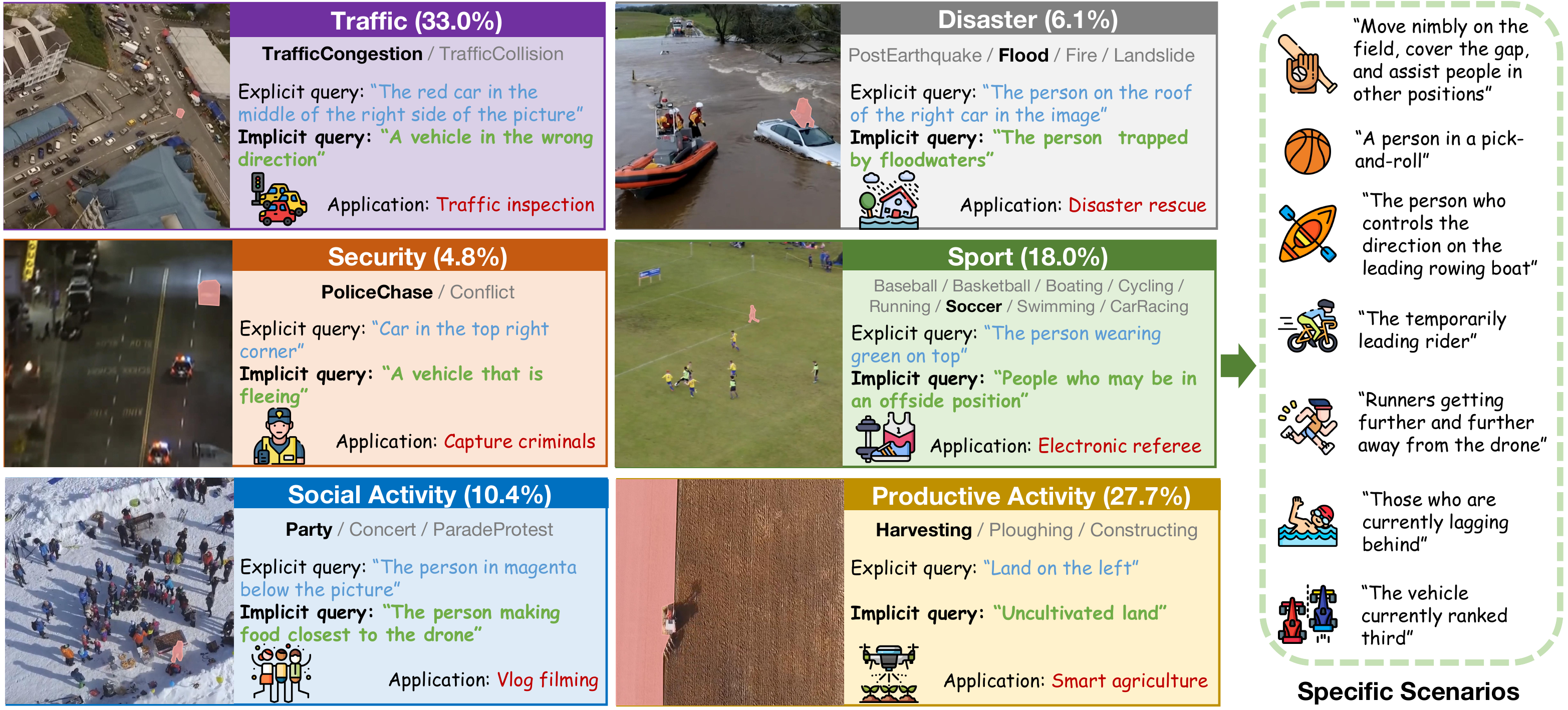}
    \caption{Visualization of the six UAV application scenarios in DVGBench. Each of the main scenarios also includes some sub-scenes. It is worth mentioning that all questions are manually labeled, rather than generated by LVLMs. Therefore, the questions cover a diverse range of knowledge points and are more challenging compared to existing RS VG datasets.
    }
    % \vspace{-0.3cm}
    \label{fig:benchmark}
\end{figure*}

\subsection{Region-level VG in RS}

Region-level VG represents object locations with bounding boxes and originates from object detection. In RS, RSVG \citep{sun2022visual} released the first region-level VG dataset and proposed GeoVG to learn numerical geospatial relations via a language encoder with a relation graph. DIOR-RSVG \citep{zhan2023rsvg} scaled to 38,320 image–query pairs and introduced a Transformer-based multi-granularity fusion module (MGVLF) leveraging multi-scale visual features and text embeddings. RSVG-HR \citep{lan2024lqvg} filled the high-resolution gap with 2,650 pairs and proposed LQVG, a language-query multimodal Transformer that retrieves targets from multi-scale features. OPT-RSVG \citep{li2024lgpa}  offered a more challenging benchmark (48,952 pairs) and a language-guided progressive visual attention framework. To handle oriented objects, GeoChat \citep{kuckreja2024geochat} provided the first rotated-box referring detection dataset; VRSBench \citep{li2024vrsbench} likewise adopts rotated annotations. In UAV imagery, AAVG \citep{zhou2024geoground} introduced the first aerial ground-vehicle referring detection dataset with 3D spatially aware expressions. However, the referring texts used in the aforementioned VG datasets are all explicit references composed of color, relative size, relative position, and object type, and do not involve the implicit references discussed in this paper.

\subsection{Pixel-level VG in RS}

Pixel-level VG in RS extends segmentation by localizing objects using masks. RefSegRS \citep{yuan2024rrsis} pioneered a generalized referring segmentation dataset (4,420 image–language–label triplets) and proposed a language-guided cross-scale enhancement module. RRSIS-D \citep{liu2024rotated} contributed 17,402 SAM-assisted triplets and introduced adaptive rotated convolution for rotation robustness, while RemoteSAM \citep{yao2025remotesam} released the largest generalized referring segmentation dataset corpus to date, utilizing an automated data engine coupled with a vision–language model to enrich semantic categories and attribute descriptions. Recently, SegEarth-R1 \citep{li2025segearth} has begun to focus on pixel-level VG tasks that require reasoning capabilities, but both are based on satellite images. Furthermore, most of them utilize large models to generate the referring text. Limited by the capabilities of existing models, the implicit references generated by these LVLMs are not challenging enough. This is also reflected in the subsequent experimental section, where mainstream models perform poorly on our manually annotated DVGBench.

\section{\dataset}

\begin{figure*}
    \centering
    \includegraphics[width=1.0\textwidth]{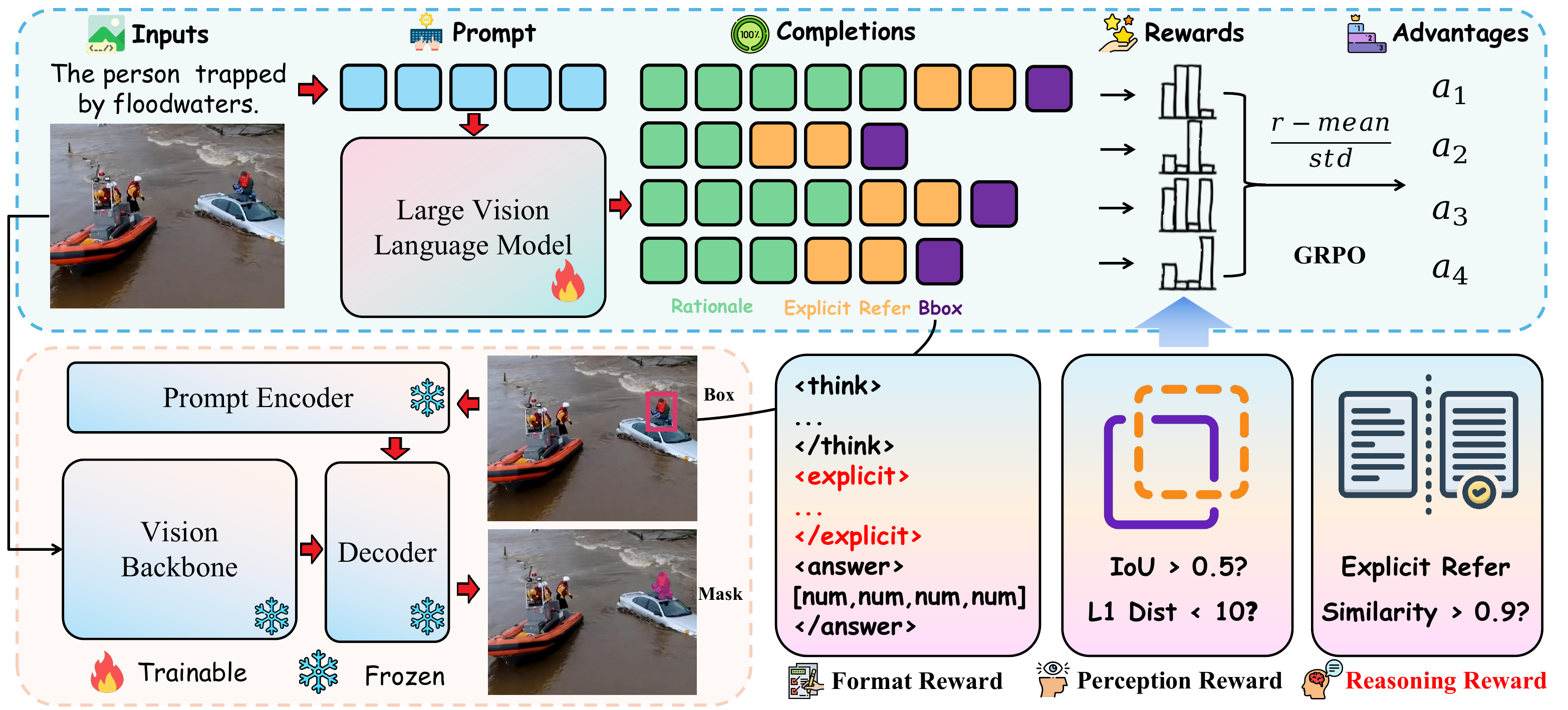}
    \caption{Framework of the DroneVG-R1, which comprises a reasoning model and a segmentation model. The reasoning model is an LVLM that generates reasoning chains and provides box-level results. Subsequently, the segmentation model produces a pixel-wise mask based on the box. In addition to regular format rewards and perceptual rewards, we have also designed a reasoning reward to enhance the quality of the model's implicit-to-explicit conversion through human-annotated explicit references.
    }
    \vspace{-0.3cm}
    \label{fig:framework}
\end{figure*}

Most existing datasets for VG are derived from object detection benchmarks~\citep{sun2022visual,zhan2023rsvg}. These datasets typically feature a limited range of object categories and exhibit significant homogeneity in referential expressions. To address these limitations, we construct our evaluation samples based on the ERA \citep{mou2020era} and VisDrone \citep{zhu2021detection} datasets. These drone datasets contain images spanning multiple diverse scenarios, ensuring a broad scope for the questions we can set.

As illustrated in Figu \ref{fig:benchmark}, DVGBench consists of UAV imagery spanning six primary application scenarios: traffic, disaster, sport, security, social activity, and productive activity. Each category comprises multiple subcategories; for example, the sport category includes eight activities—baseball, basketball, boating, cycling, running, soccer, swimming, and car racing. For example, it can be used to determine whether a vehicle has violated traffic rules by analyzing traffic signs and the vehicle’s orientation, to locate fugitives pursued by the police by tracking searchlights projected onto the ground by helicopters at night, and to assess whether an offensive player is offside from an overhead view of the soccer field. Implicit reference based on drone imagery has broad application value in such contexts. We also add the proportion of each category in Figure \ref{fig:benchmark}. The results indicate that the traffic category constitutes the largest proportion at 33.0\%, while the disaster and sociaty security categories account for relatively smaller proportions, at 6.1\% and 4.8\%, respectively. The uneven distribution of question types across scenarios in DVGBench reflects a realistic data collection bias. To ensure a fair evaluation, we employ a category-averaging method when calculating the overall accuracy scores.

However, constructing implicit references presents greater challenges than explicit ones, as each question must incorporate scenario-specific commonsense knowledge, making it impossible to adopt template-based generation or LVLM-assisted creation with human verification. To ensure benchmark reliability, we employ a fully manual annotation process in which every scenario-specific question was carefully crafted using domain knowledge to authentically reflect real-world requirements. Notably, our annotation revealed that in most scenarios, humans instinctively combine referential expressions with contextual knowledge rather than relying on explicit cues such as visual attributes or spatial relationships, demonstrating that implicit references better align with practical user needs. We annotated a total of 2,863 samples. Each question contains two types of queries: explicit and implicit. The dataset is split into a test set of 873 samples and a training set of 1,990 samples. Compared to existing VG datasets, DVGBench boasts the following three advantages and characteristics:

% \subsection{Scene-Specific Expertise}
% \label{sec:SSE}
\textbf{Scene-Specific Expertise.} 
Scene-specific domain knowledge is embedded during the construction of implicit referring expressions. The benchmark spans six UAV application domains, each encoding the specialized knowledge pertinent to its context: (1) Traffic—reasoning about traffic violations, lane-change intent, and liability attribution in accidents, with the objective of assisting traffic police in monitoring traffic violations and handling traffic incidents; (2) Disaster—identification of individuals awaiting rescue and delineation of safe zones, to facilitate rapid rescue of trapped persons; (3) Security—recognition of illegal vehicles and police and patrol cars, as well as demonstrators and police officers, to support fugitive apprehension and maintenance of public order; (4) Sport—soccer offside determination, race ranking identification, and detection of pick-and-roll participants in basketball, enabling intelligent commentary or refereeing and complementing ground-based camera viewpoints; (5) Social Activity—hazard detection and identification of designated individuals, to ensure event safety; and (6) Productive Activity—discrimination between cultivated and uncultivated parcels, enabling real-time monitoring of agricultural production progress.

\textbf{Explicit-Implicit Pair.} 
An Implicit–Explicit Pair (IEP) evaluation protocol is introduced in DVGBench, where each referent is associated with both explicit and implicit referring expressions, enabling: (1) controllable, layered assessment via an explicit-to-implicit difficulty gradient from perception to scene reasoning; (2) quantification of the reasoning gap as the explicit–implicit localization performance differential; and (3) consistency verification by measuring agreement between explicit and implicit localizations with a paired-consistency metric to assess robustness to paraphrase and expression variability.

% \subsection{Annotation Pipeline and Quality Control}
% \label{sec:AQC}
\textbf{Annotation Pipeline and Quality Control.} 
The DVGBench annotation pipeline comprises three stages—implicit reference annotation, explicit reference annotation, and mask annotation—and incorporates safeguards to ensure reliability. First, for implicit references, images were curated from two UAV datasets with diverse aerial scenes, ERA and VisDrone, and a subset satisfying the conditions for implicit referring-expression annotation was manually labeled. Second, explicit references for the same targets were constructed using only explicit, scene-agnostic attributes such as color and relative spatial position, deliberately avoiding any reliance on domain knowledge. Third, target contours were delineated with polygonal annotations to produce the segmentation masks. 

To minimize annotator-induced bias, we developed comprehensive and explicit annotation guidelines. All annotators underwent training to ensure consistent application of the criteria for determining ground-truth positions. The annotation procedure consisted of multiple stages: an initial annotation was performed by one annotator, followed by a validation check by a second, senior annotator. Cases with disagreements were reviewed and resolved through discussion with a third expert. Instances that remained unresolved after this process were excluded from the final dataset. Furthermore, to assess annotation reliability, 10\% of the items were independently re-annotated. Inter-annotator agreement was quantified using Cohen’s Kappa, and an average consistency score of above 0.95 was maintained.

% To ensure quality, a cross-annotation workflow was adopted: Annotator A wrote both implicit and explicit texts; Annotator B produced masks conditioned on these texts and revised problematic expressions, which were verified by Annotator C; unresolvable items were discarded. Additionally, 10\% of items underwent independent mask re-annotation, with agreement measured by Cohen’s Kappa and an average consistency coefficient enforced above 0.95.

% \section{\dataset}

\section{Method}

The framework of the proposed model is illustrated in Figure~\ref{fig:framework}. Named DroneVG-R1, our method decouples the reasoning and segmentation processes, yielding significant improvements in both flexibility and performance. By integrating the robust reasoning ability of GRPO with I2E-CoT and the precise segmentation capability of SAM~\citep{kirillov2023segment}, our approach effectively tackles the challenge of implicit VG in drone imagery.

\subsection{Framework of DroneVG-R1} 

DroneVG-R1 adopts a two-stage strategy comprising an LVLM model and a segmentation model~\citep{liu2025seg}. The LVLM consists of three core components: a vision encoder, a multimodal adaptor, and an LLM backbone. The vision encoder utilizes a native dynamic-resolution Vision Transformer (ViT). Its key innovation is the incorporation of a windowed attention mechanism, which effectively decouples computational complexity from input resolution. The multimodal adaptor is designed as a two-layer Multilayer Perceptron. Its primary function is to compress and project the high-dimensional image features from the vision encoder into the text embedding space of the language model. The LLM backbone is initialized using the powerful, extensively pre-trained Qwen2.5 language model. Specifically, given an image $\mathbf{I}$ and a high-level text instruction $\mathbf{T}$, DroneVG-R1 $\mathcal{F}$ generates an interpretable reasoning process and subsequently produces the expected output corresponding to $\mathbf{T}$. The model output is represented in a structured format, from which we extract the bounding boxes $\mathbf{B}$ and center point $\mathbf{P}$ to serve as input to segmentation models. This process can be formulated as follows:
\begin{equation}
     (\{\mathbf{B}_i, \mathbf{P}_i\})_{i=1}^{N} = \mathcal{F}(\mathbf{I}, \mathbf{T}).
\end{equation}
Subsequently, the masks $\mathbf{A}$ are predicted by the segmentation model $\mathcal{M}$ using the extracted bounding boxes $\mathbf{B}$ and center point $\mathbf{P}$:
\begin{equation}
     \mathbf{A}_i = \mathcal{M}(\mathbf{B}_i, \mathbf{P}_i).
\end{equation}

In fact, the segmentation model is not our primary focus, as it can be conveniently replaced without requiring training. We experimented with various SAM-like~\citep{kirillov2023segment} models and ultimately selected SAM2 as the default segmentation model. Subsequent experiments demonstrate that although DroneVG-R1 cannot natively output pixel-level results, its powerful reasoning capability, with the aid of tools, enables it to significantly outperform existing models on our DVGBench. The advantage of this two-stage framework lies in decoupling reasoning and segmentation abilities, allowing for independent optimization. Should more advanced SAM models emerge in the future, the segmentation performance of DroneVG-R1 can be further enhanced.

% \subsection{I2E-CoT} 

\subsection{GRPO with I2E-CoT}

\begin{figure}
    \centering
    \includegraphics[width=1.0\linewidth]{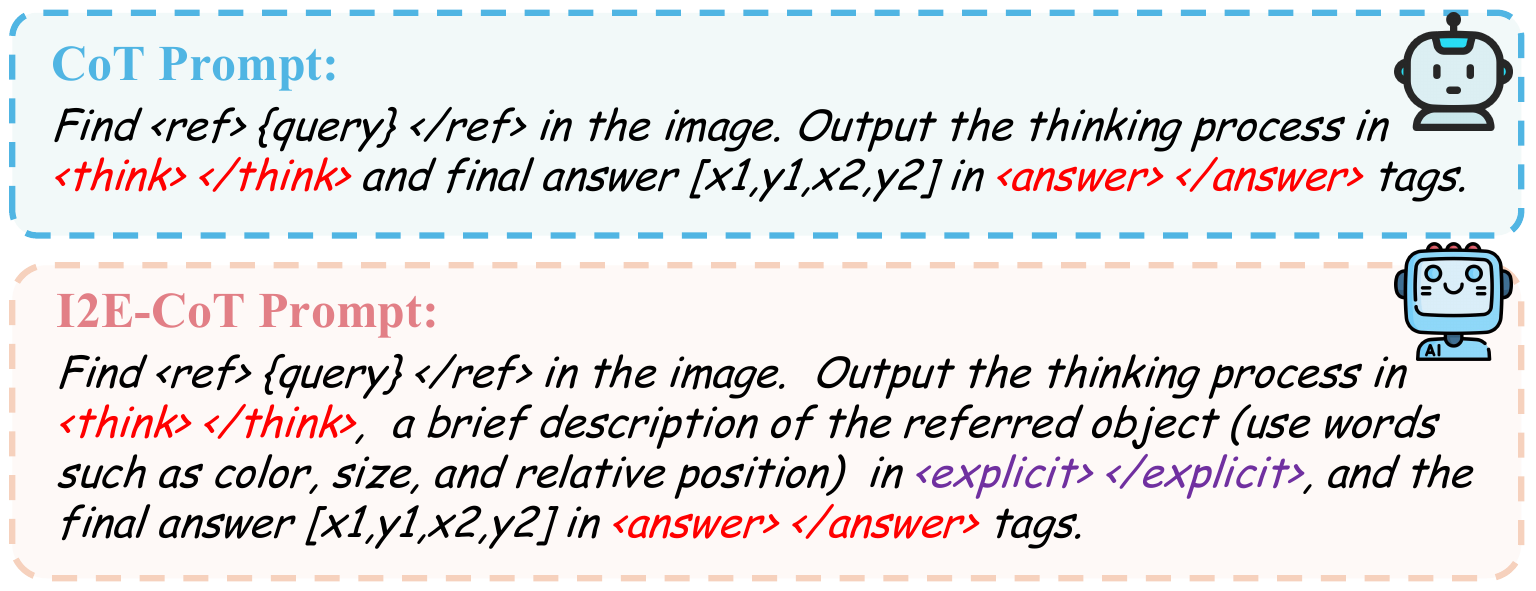}
    \caption{Prompt templates of CoT and I2E-CoT. 
    }
    \label{fig:prompt}
\end{figure}

As illustrated in Figure~\ref{fig:prompt}, the prompt template of the proposed I2E-CoT features a significant modification. In addition to the thinking process and the final answer, it explicitly mandates the model to produce a brief description of the referred object. This description, enclosed within new \texttt{<explicit> </explicit>} tags, must utilize perceptual attributes such as color, size, and relative position. This design strategically forces the model to translate an implicit query into an explicit, grounded description before committing to a final bounding box, thereby enhancing the reliability of the reasoning process.

Using \texttt{<explicit>} to mark the explicit reference corresponding to the current implicit reference, we apply reinforcement learning to guide the model in consistently triggering I2E conversion behavior. Let $\mathcal D_{RFT}=\{(I, q_\mathrm{i}, q_\mathrm{e}, a)\}$ denote the dataset annotated with explicit and implicit references, where $q_\mathrm{i}$ denotes the implicit reference and points to the same object as the explicit reference $q_\mathrm{e}$. Let $y$ denote a response sampled from the policy $\pi_\theta(\cdot\mid x)$. Given a reward function $r(I, q_\mathrm{i}, q_\mathrm{e}, a)\in\mathbb R$, the policy objective is
\begin{equation}
  \mathcal{L}_{\mathrm{RFT}}(\theta)
  \;=\;
  \mathbb{E}_{x\sim\mathcal D_{RFT},\;y\sim\pi_\theta(\cdot\mid x)}
  \bigl[R(x, q_\mathrm{i}, q_\mathrm{e}, a)\bigr].
  \label{eq:rl-sentence-obj}
\end{equation}
where $x = [I; q_\mathrm{i}]$ concatenates the image and implicit reference, $y = [q_\mathrm{e}; a]$ concatenates the explicit reference and answer sequence, and $R$ is the reward function.

Unlike reinforcement learning algorithms such as PPO ~\citep{gu2021proximal}, which require an additional critic model to estimate policy performance, GRPO directly compares groups of candidate responses, thereby eliminating the need for a separate critic network. Given a query $q$, It samples $N$ candidate responses $\{y_1, y_2, \ldots, y_N\}$ from the policy $\pi_\theta$ and evaluates each response $y_i$ using the reward function $R$. To determine the relative quality of these responses, GRPO normalizes the rewards by computing their mean and standard deviation, and subsequently derives the advantage as:
\begin{equation}
    A_i = \frac{r_i - \text{mean}\{r_1, r_2, \ldots, r_N\}}{\text{std}\{r_1, r_2, \ldots, r_N\}},
\end{equation}
where $A_i$ represents the advantage of the candidate response $y_i$ relative to other responses sampled within the group. GRPO encourages the model to generate responses with higher advantages by optimizing the policy $\pi_\theta$ through the following objective:
\begin{equation}
\begin{split}
    \mathcal{J}_{GRPO}&(\theta) = \mathbb{E}[{\{y_i\}_{i=1}^N\sim\pi_{\theta_{old}}(q)}] \\
    &\frac{\sum_{i=1}^N\left\{\min[s_1 A_i,\ s_2  A_i]-\beta\mathbb{D}_{KL}[\pi_\theta||\pi_{ref}]\right\}}{N} \\
    s_1 & = \frac{\pi_\theta(y_i|q)}{\pi_{\theta_{old}}(y_i|q)}; \quad s_2  = \text{clip}\left(s_1,1+\epsilon,1-\epsilon\right)
\end{split}
\end{equation}
where $\epsilon$ and $\beta$ are the clipping hyperparameters and the KL divergence penalty coefficient, respectively. The expectation term, $\mathbb{E}[{\{y_i\}_{i=1}^N\sim\pi_{\theta_{old}}(q)}]$, indicates that the objective is averaged over all possible groups of $N$ responses sampled from the old policy $\pi_{\theta_{old}}(q)$. The policy improvement component, $\min[s_1 A_i,\ s_2  A_i]$, stably guides the policy to favor high-advantage responses by leveraging their relative quality within the group and a clipping mechanism, where $\epsilon$ is the clipping threshold. The regularization term, $-\beta\mathbb{D}_{KL}[\pi_\theta||\pi_{ref}]$, ensures training stability by penalizing large deviations from the reference policy, thereby preserving response quality and diversity. The optimization process follows a clear iterative cycle: The process begins with sampling a group of $N$ responses for a given prompt from the current old policy. Subsequently, the advantage score $A_i$ for each response is calculated, typically by a reward model. On this basis, the system quantifies the performance of the entire group of responses using the objective function formula, and thereby updates the model parameters via gradient ascent to maximize the expected advantage. This cycle repeats continuously, with the updated new policy becoming the old policy for the next round of sampling, progressively enhancing the model's performance.

\subsection{Reward Functions}

A sophisticated reward system was designed, incorporating \textit{format}, \textit{perception}, and \textit{reasoning} rewards, to better guide the optimization of I2E reasoning.

\textbf{Format Reward.} We utilize the format reward to ensure the model's response strictly adheres to the required format. It can be divided into two parts: \textbf{1) Overall Format Reward:} To force the model to think deeply before answering, we add the format \texttt{<think>} Thinking Process Here \texttt{</think> <explicit>} Explicit Referrence Here \texttt{</explicit> <answer>} Final Answer Here \texttt{</answer>} to constrain the model; \textbf{2) Box Format Reward:} To make the bounding box format predicted by the model processable by regular expressions, we also add \texttt{[Number, Number, Number, Number]}. The reward is 1 when the regular expression meets the requirement; otherwise, it is 0.

\textbf{Perception Reward.} To help the model ground the referred area, we utilize the perception reward, which mainly contains: \textbf{1) IoU Reward:} We calculate the Intersection over Union (IoU) between output bounding boxes and ground truth bounding boxes. If $\text{IoU} > 0.5$, the reward is 1; otherwise, the reward is 0.  \textbf{2) L1 Reward:} This reward evaluates the L1 distance between the predicted bbox and the ground-truth bbox. A
reward of 1 is assigned if their L1 distance less than 10 pixels; otherwise, the reward is 0. IoU and L1 loss offer continuous, geometry-aware supervision, demonstrating robustness against variations in text formatting. 

% \textbf{2) L1 Reward:} We compute the L1 distance between output and ground truth bounding boxes (including points). If the $\text{L1 distance} < 10$, the reward is 1; otherwise, the reward is 0;

\textbf{Reasoning Reward.}
As the saying goes, \textit{``To perceive is to see what is, but to reason is to understand why''}, the transition from implicit to explicit references demands not only perceptual recognition but also deeper reasoning. Specifically, we compute the Jaccard text similarity \citep{jaccard1901etude} between the model-generated explicit reference and the ground-truth explicit expression. A reward of 1 is assigned if $\text{similarity} > 0.9$; otherwise, the reward is 0. The reasoning reward is designed to leverage the human-annotated explicit references in DVGBench to guide the model in learning the correct implicit-explicit conversion patterns.

The above three sets of reward functions collectively support the reinforcement fine-tuning of DroneVG-R1. The reward functions of our framework are formulated as a linear combination of three components: format reward, perception reward, and reasoning reward. The experimental results demonstrate that the reasoning reward is even more critical than the perception reward.

\subsection{Limitation}

While the I2E-CoT method offers significant benefits, its limitations must be considered. The method's effectiveness is directly contingent upon the LVLM's capacity for generating valid intermediate reasoning steps. As a result, it is not suitable for models designed solely for direct output without explanatory capabilities, or for LVLMs that lack reasoning transparency. This scope limitation means our approach is primarily relevant to—and has been validated on—LVLMs with inherent reasoning capabilities. Future work could explore ways to distill reasoning steps or adapt the method for a broader class of models.

\begin{table*}
\caption{Performance (Acc@0.5\%) comparison of region-level VG with implicit queries on DVGBench. The notation ``SFT" denotes supervised fine-tuning. Since our dataset lacks annotations for CoT, Qwen2.5-VL loses the ability to output reasoning processes after sft. DroneVG-R1 retains Qwen2.5-VL's ability to output reasoning processes thanks to the use of GRPO.} 
\setlength{\tabcolsep}{3pt}
\resizebox{\textwidth}{!}{
\begin{tabular}{lcccccccc}
\toprule
 Model    & Rationale & Security & Traffic & Social Activity & Disaster & Productive Activity & Sport & AVG \\ \midrule
\multicolumn{9}{l}{\hfill \textit{Models w/o reasoning processes} } \\
\midrule
MGVLF \citep{zhan2023rsvg} & \icono  & 7.14\%  & 0.00\%  & 13.19\%  & 5.66\% &4.96\%  & 3.18\%  & 5.69\%   \\
LQVG \citep{lan2024lqvg} & \icono  & 2.38\%  & 13.54\% & 9.89\%  & 1.89\%   & 6.20\%  & 8.28\% & 7.03\%   \\
GeoChat \citep{kuckreja2024geochat} & \icono  & 11.91\%  & 2.43\% & 8.79\%  & 9.43\%   & 4.55\%  & 3.82\% & 6.82\%   \\
LHRS-Bot \citep{muhtar2024lhrs} & \icono  & 0.00\%  & 1.04\% & 2.20\%  & 0.00\%   & 2.07\%  & 5.10\% & 1.73\%   \\
GeoGround \citep{zhou2024geoground} & \icono  & 19.05\%   & 18.06\%  & 29.67\%  & 20.75\%    & 33.88\%   & 11.46\%  & 22.15\%    \\
MGVLF (SFT) &  \icono  &  4.76\%  & 0.35\% & 5.49\%  & 0.00\%   & 20.66\%  & 1.91\% & 5.53\%   \\
LQVG (SFT)  &  \icono  & 33.33\%  & 23.61\% & 16.48\%  & 15.09\%   & 43.39\%  & 19.75\% & 25.28\%   \\
Qwen2.5-VL 3B (SFT) \citep{bai2025qwen2} & \icono  & 45.24\%  & 20.14\% & 41.76\%  & \textbf{47.17\%}   & 45.45\%  & 40.13\% & 39.98\%  \\
Qwen2.5-VL 7B (SFT) & \icono  & 54.76\%  & 48.61\% & \textbf{45.05\%}  & 41.51\%   & 66.53\%  & \textbf{45.22\%} & 50.28\%   \\
\midrule
\multicolumn{9}{l}{\hfill \textit{Models w/ reasoning processes} } \\
\midrule
DeepSeek-VL2-Tiny \citep{wu2024deepseek} & \icoyes  & 21.43\%   & 8.33\%  & 16.48\%   & 15.09\%    & 31.82\%   & 24.20\%  & 19.56\%    \\
% DeepSeek-VL2-Small  & \icono  & -  & - & -  & -   & -  & - & -   \\
InternVL3.5 4B \citep{chen2024internvl} & \icoyes  & 42.86\%  & 14.58\% & 14.29\%  & 16.98\%   & 17.36\%  & 18.47\% & 20.76\%   \\
InternVL3.5 8B & \icoyes  & 33.33\%   & 28.47\%  & 27.47\%   & 26.42\%    & 31.40\%   & 25.48\%  & 28.76\%    \\
Qwen2.5-VL 3B   & \icoyes  & 45.24\%  & 25.69\% & 29.67\%  & 22.64\%   & 51.24\%  & 35.03\% & 34.92\%   \\
Qwen2.5-VL 7B  &  \icoyes  & 35.71\%  & 41.32\% & 34.07\%  &  \textbf{47.17\%}   & 63.64\%  & 40.76\% & 43.78\%   \\
\rowcolor{mylightblue}DroneVG-R1 3B &  \icoyes  & 50.00\%  & 42.36\% & 41.76\%  & 30.19\%   & 57.85\%  & 40.13\% & 43.71\%   \\
\rowcolor{myblue}DroneVG-R1 7B & \icoyes   & \textbf{57.14\%}  & \textbf{52.08\%}  & \textbf{45.05\%}  & 43.40\%  & \textbf{70.66\%}  & 43.95\%  & \textbf{52.05\%}  \\

\bottomrule
\end{tabular}}
\label{tab:rec_overall}
\end{table*}

\section{Experiments}

Our experimental setup largely follows GeoGround \citep{zhou2024geoground}. We adopt the GRPO algorithm \citep{wu2024deepseek} with an initial learning rate of 1e-6, which is warmed up with a ratio of 0.01 before applying a linear decay schedule. The number of GRPO generations is set to 8. To alleviate GPU memory usage, all models are fine-tuned using LoRA with a rank of 64, combined with ZeRO-2 stage memory optimization. Given that Qwen2.5-VL already exhibits strong visual grounding capabilities, we fine-tune it for only 1 epoch. All models are trained on 8 NVIDIA RTX 4090 GPUs (24GB each), with a global batch size of 32. We utilize the asynchronous mode of GRPO provided by the ms-swift \citep{zhao2024swift} framework to decouple training and inference resources. Specifically, 2 GPUs are allocated for rollout, and the remaining 6 GPUs are used for training. The inference batch size is set to 1 across all experiments. 

% \subsection{Experimental Setup}

\subsection{Region-level VG}

\textbf{Settings.} 
We follow standard evaluation protocols~\citep{pang2025vhm,li2024vrsbench} and assess the region-level VG task using the Acc@0.5 metric, which measures the proportion of predicted HBBs with an IoU greater than 0.5 relative to the ground truth boxes. For the GeoChat~\citep{kuckreja2024geochat}, we convert its output OBBs to HBBs.

\textbf{Results on DVGBench.} The performance of DroneVG-R1 is compared against two specialized models and eight mainstream LVLMs on DVGBench in Table~\ref{tab:rec_overall}. Notably, half of these models (5 out of 10) can generate reasoning processes, while the other half are limited to outputting bounding boxes. A key finding is the superior performance of our proposed DroneVG-R1 model, which integrates the proposed I2E-CoT framework. The DroneVG-R1 7B model achieves a state-of-the-art average accuracy of 52.05\%, significantly outperforming all other benchmarks. This represents a substantial improvement of over percentage points compared to its base model, Qwen2.5-VL 7B (43.78\%), when both are evaluated with reasoning processes enabled. This performance gain is consistent across most semantic categories (e.g., Security, Traffic, Productive Activity), demonstrating the efficacy of I2E-CoT in aligning the reasoning process with accurate region prediction. Notably, DroneVG-R1 7B also surpasses the strong baseline of Qwen2.5-VL 7B (SFT)—which does not output rationales—by nearly 2 percentage points (50.28\% vs. 52.05\%). This is particularly significant as it indicates that our method not only preserves but enhances model performance while retaining the interpretable reasoning capability, a common trade-off in SFT. As noted, standard SFT on our dataset causes Qwen2.5-VL to lose its reasoning ability, a limitation circumvented by I2E-CoT.

\begin{table*}
\caption{Performance (Acc@0.5\%) comparison of region-level VG with both implicit and explicit queries on the strict version of DVGBench (DVGBench-hard). Consistency can measure whether the bounding boxes output by a model for the explicit reference and implicit reference of the same object are consistent.} 
\setlength{\tabcolsep}{3pt}
\resizebox{\textwidth}{!}{
\begin{tabular}{l|cccccc|c|c}
\toprule
 Model     & Security & Traffic & Social Activity & Disaster & Productive Activity & Sport & AVG & Consistency\\ \midrule
GeoGround \citep{zhou2024geoground}  & 16.67\%& 9.38\% & 4.40\%  & 15.09\%   & 13.64\%  & 7.01\% & 11.03\% & 3.60\%   \\
DeepSeek-VL2-Tiny \citep{wu2024deepseek}   & 16.67\%  & 6.25\% & 13.19\%  & 11.32\%   & 26.45\%  & 21.66\% & 15.92\%  & 29.00\%  \\
% DeepSeek-VL2-Small  & \icono  & -  & - & -  & -   & -  & - & -   \\
InternVL3.5 4B \citep{chen2024internvl}   & 21.43\%  & 2.78\% & 2.20\%  & 7.55\%   & 4.96\%  & 6.37\% & 7.55\% & 3.70\%  \\
InternVL3.5 8B  & 9.52\%  & 10.07\% & 13.19\%  & 9.43\%   & 11.57\%  & 8.28\% & 10.34\%  & 4.10\% \\
Qwen2.5-VL 3B  \citep{bai2025qwen2}   & 38.10\%  & 14.58\% & 17.58\%  & 11.32\%   & 33.06\%  & 26.75\% & 23.57\% & 31.30\%  \\
Qwen2.5-VL 7B    & 26.19\%  & 29.17\% & 25.27\%  & 37.74\%   & 52.89\%  & 35.03\% & 34.38\%  & 44.70\% \\

\rowcolor{mylightblue}DroneVG-R1 3B   & 45.24\%  & 32.64\% & 28.57\%  & 22.64\%   & 41.74\%  & 31.21\% & 33.67\%  & 33.40\% \\
\rowcolor{myblue}DroneVG-R1 7B    & \textbf{50.00\%}  & \textbf{41.67\%}  &  \textbf{38.46\%}  &  \textbf{41.51\%}  & \textbf{61.57\%}  & \textbf{40.76\%}  & \textbf{45.66\%} & \textbf{47.60\%} \\

\bottomrule
\end{tabular}}
\label{tab:rec_overall_strict}
\end{table*}

\begin{figure*}
    \centering
    \includegraphics[width=1.0\textwidth]{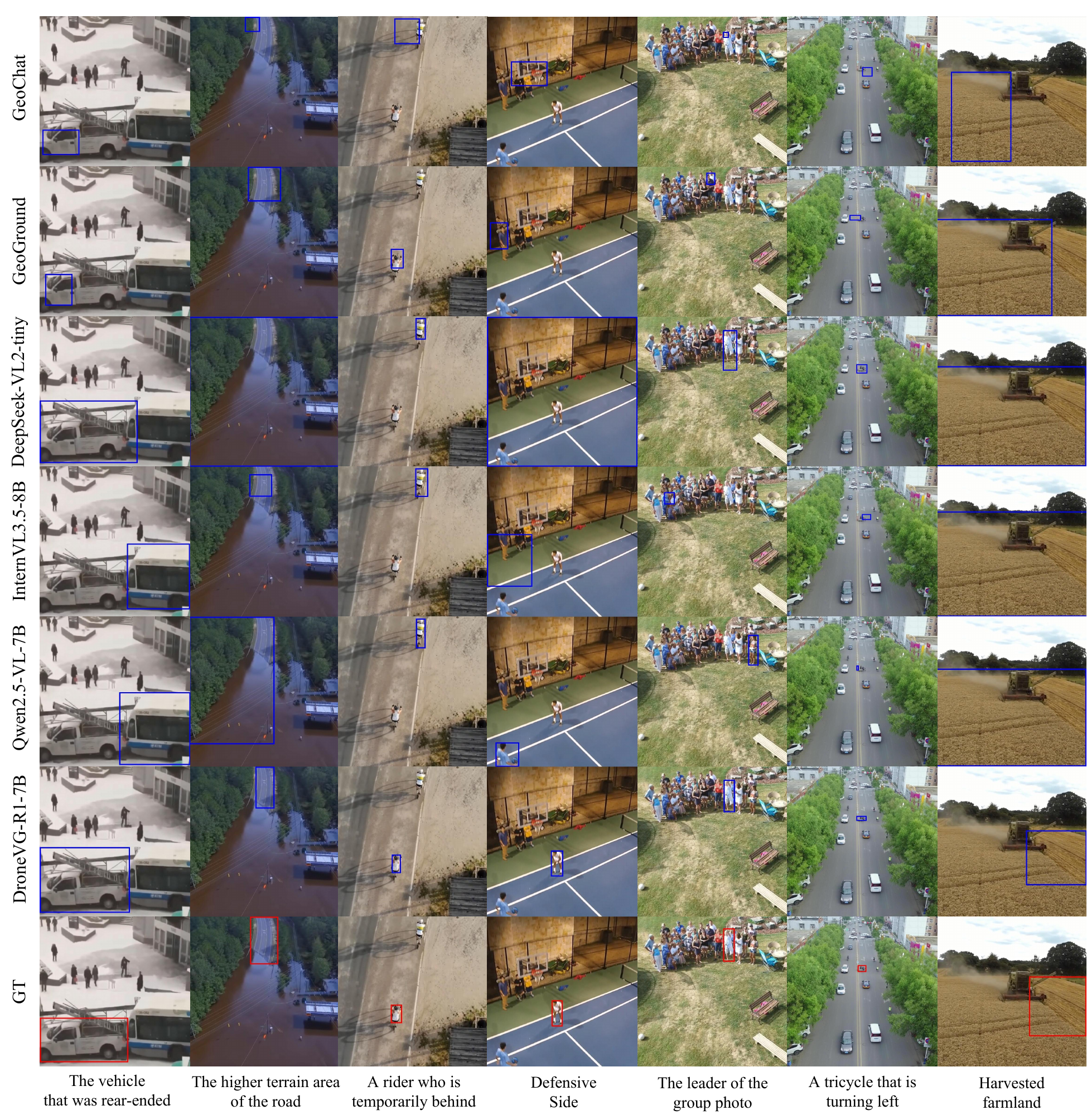}
    \caption{Visualizations of DroneVG-R1 and other LVLMs on DVGBench. Blue bounding boxes represent prediction, while red boxes indicate ground truth.
    }
    \label{fig:viz_box}
    \vspace{-0.5cm}
\end{figure*}

Furthermore, the results underscore the critical impact of high-quality rationale generation. While several models (e.g., InternVL3.5 8B, DeepSeek-VL2-Tiny) are capable of producing reasoning chains, their average accuracies (28.76\% and 19.56\%, respectively) lag considerably. This suggests that the mere presence of a rationale is insufficient; its quality and relevance, which are directly optimized by I2E-CoT, are paramount for accurate grounding. Additionally, it should be noted that even after fine-tuning the training set provided with DVGBench, the two specific models MGVLF and LQVG, which do not incorporate LLMs, showed only limited improvement. This is attributed to the high difficulty of the DVGBench samples and the significant variation in manually annotated question types, which make it challenging for the specific models to achieve a good fit. Finally, the scaling effect is confirmed within our own architecture, as the DroneVG-R1 7B variant consistently outperforms its 3B counterpart (52.05\% vs. 43.71\% AVG), validating the benefit of increased model capacity.

\textbf{Results on DVGBench-hard.} The performance of region-level VG models is further evaluated on a more challenging strict version of the benchmark, DVGBench-hard, with results detailed in Table~\ref{tab:rec_overall_strict}. DVGBench evaluates only implicit queries, considering a prediction correct if the IoU between the implicit query and the ground-truth bounding box exceeds 0.5. In contrast, DVGBench-hard evaluates both explicit and implicit queries, requiring the IoU of both prediction types with the ground-truth box to be greater than 0.5 for a correct result. This evaluation introduces a critical new metric, Consistency, which measures whether a model's bounding box predictions for explicit and implicit references to the same object are aligned, thereby assessing the stability of its understanding. Table~\ref{tab:rec_overall} only evaluates the Acc@0.5 corresponding to implicit queries, making it impossible to compute the consistency metric.

On this demanding dataset, the proposed DroneVG-R1 7B model demonstrates superior robustness and achieves state-of-the-art performance. It attains the highest average accuracy of 45.66\%, significantly outperforming all comparative models. This represents a substantial improvement of over 11 percentage points compared to its base model, Qwen2.5-VL 7B (34.38\%), underscoring the efficacy of the proposed method under challenging conditions. The performance advantage is consistent across all six categories, with DroneVG-R1 7B leading in each, most notably in Productive Activity (61.57\%) and Security (50.00\%).

A key observation is the model's exceptional performance on the Consistency metric. DroneVG-R1 7B achieves a consistency score of 47.60\%, which is the highest among all models and closely aligns with its overall accuracy. This high level of agreement indicates that the model's predictions are not merely coincidental but are based on a stable and coherent understanding of the target objects, regardless of how they are referred to in the query. In contrast, while the base model Qwen2.5-VL 7B also shows reasonable consistency (44.70\%), its lower accuracy suggests a stable but less precise understanding. Other models, such as InternVL3.5 8B, exhibit a dramatic disconnect between their low accuracy (10.34\%) and an even lower consistency (4.10\%), implying that their occasional correct predictions may be unreliable and not grounded in a consistent perceptual reasoning process. Furthermore, the scaling effect is validated, as the DroneVG-R1 7B variant outperforms its 3B counterpart (33.67\% AVG, 33.40\% Consistency). Notably, the proposed method enables the smaller DroneVG-R1 3B model to compete closely with, and even surpass in some categories, the larger 7B parameter base model (Qwen2.5-VL 7B), highlighting the efficiency gains afforded by our approach.

In summary, the results on DVGBench-hard confirm that the proposed method not only achieves the highest accuracy but also ensures the most consistent and reliable model behavior. The high consistency score strongly suggests that the model performs genuine visual reasoning rather than relying on spurious correlations, marking a significant step towards building trustworthy VG systems.

\textbf{Visualization.} Figure~\ref{fig:viz_box} provides a qualitative comparison of region-level predictions generated by DroneVG-R1-7B and several competing LVLMs on diverse and challenging scenarios from DVGBench. The GT bounding boxes are annotated in red, while the model predictions are shown in blue. It can be observed that DroneVG-R1 demonstrates exceptional grounding accuracy in handling drone implicit VG tasks across various scenarios. In contrast, the comparative models exhibit characteristic failure modes.

\begin{table}
\caption{Performance comparison of pixel-level VG with implicit queries on DVGBench. All three metrics are based on the masks.} 
\setlength{\tabcolsep}{2pt}
\resizebox{\linewidth}{!}{
\begin{tabular}{lcccc}
\toprule
 Model    & Rationale & mIoU & oIoU
 & Acc@0.5 \\ \midrule

RMSIN
 \citep{liu2024rotated} & \icono & 9.07\%  & 4.34\%  & 7.90\%   \\
LISA
 \citep{lai2024lisa} & \icono & 5.08\%  & 5.17\%  & 4.25\%   \\
PixelLM
 \citep{ren2024pixellm} & \icono & 27.77\%  & 38.84\%  & 27.00\%   \\
NExT-Chat
 \citep{zhang2023next} & \icono & 19.26\%  & 22.35\%  & 21.42\%   \\
GeoGround
 \citep{zhou2024geoground} & \icono & 24.57\%  & 28.92\%  & 24.94\%   \\
SegEarth-R1
 \citep{li2025segearth} & \icono & 3.89\%  & 6.02\%  & 2.75\%   \\
RemoteSAM
 \citep{yao2025remotesam} & \icono & 12.02\%  & 12.85\%  & 9.97\%   \\
GeoPix
 \citep{ou2025geopix} & \icoyes & 10.61\%  & 19.18\%  & 6.20\%   \\
GeoPixel
 \citep{shabbir2025geopixel} & \icoyes & 29.63\%  & 26.42\%  & 29.94\%   \\
\rowcolor{mylightblue}DroneVG-R1 3B & \icoyes & 38.62\%  & 36.31\%  & 42.76\%   \\
\rowcolor{myblue}DroneVG-R1 7B & \icoyes & \textbf{46.70\%}  & \textbf{44.15\%}  & \textbf{51.78\%}   \\

\bottomrule
\end{tabular}}
\label{tab:res_overall}
\end{table}

\begin{figure*}
    \centering
    \includegraphics[width=1.0\textwidth]{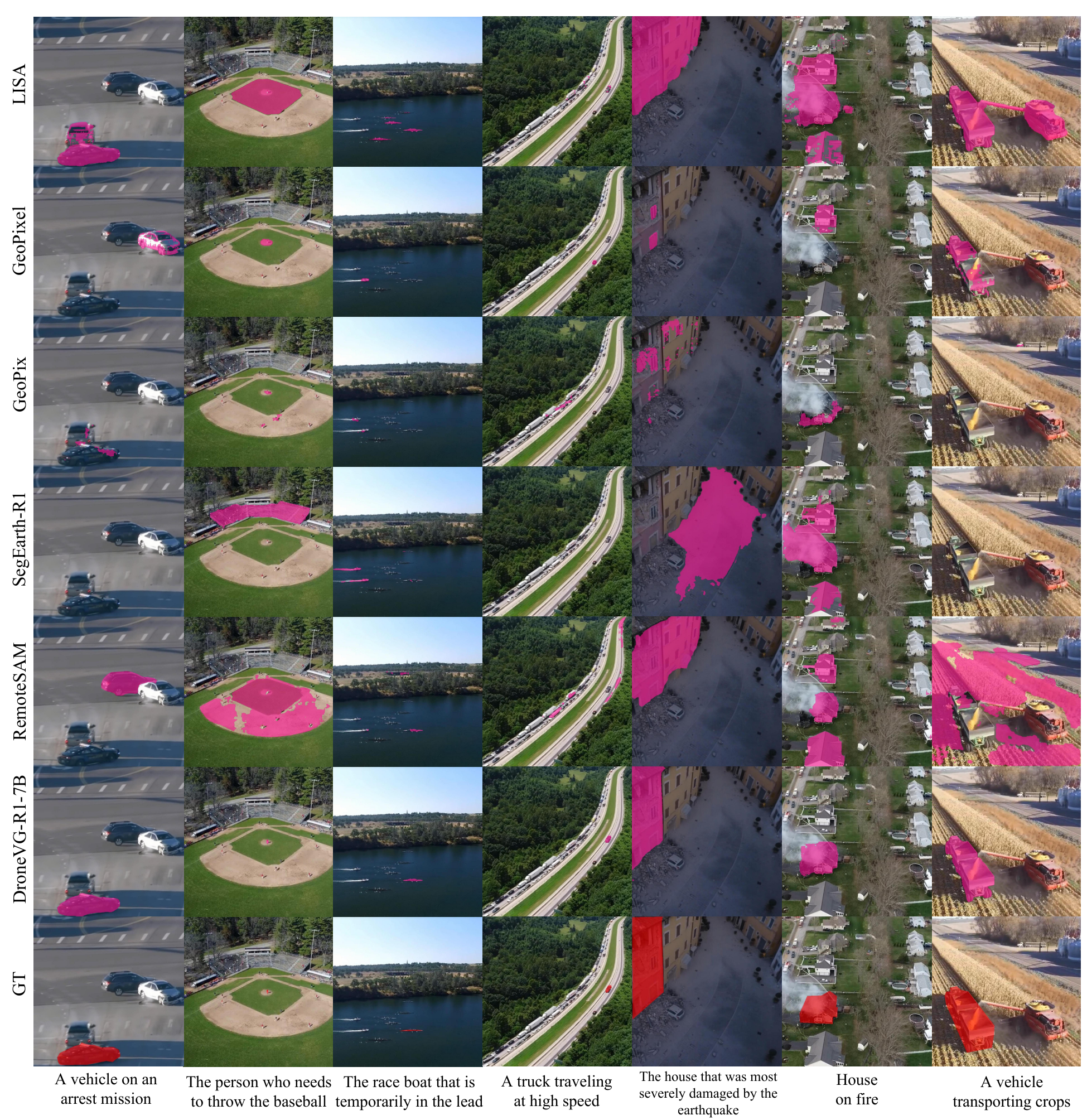}
    \caption{Visualizations of DroneVG-R1 and other LVLMs on DVGBench. Mask regions are represented using pink-colored overlays.
    }
    \label{fig:viz_seg}
    \vspace{-0.5cm}
\end{figure*}

\subsection{Pixel-level VG}

\textbf{Settings.} 
We further evaluate the models on the more challenging task of pixel-level VG with implicit queries on the DVGBench benchmark. We utilize Acc@0.5, Mean Intersection-over-Union (mIoU), and Overall Intersection-over-Union (oIoU) as evaluation metrics, similar to prior studies~\citep{wu2020phrasecut,yuan2024rrsis}. This task demands not only a high-level semantic understanding of implicit descriptions but also precise pixel-wise segmentation capabilities. 

The pixel-level VG experiment is not intended to evaluate SAM2's inherent accuracy, but to indirectly assess the quality of the spatial priors (bounding boxes) generated by DroneVG-R1. The core hypothesis is that a high-quality bounding box, which accurately captures the referred object's location, will enable a powerful segmenter like SAM2 to produce a precise mask. Consequently, an improvement in mIoU primarily reflects the spatial accuracy and effectiveness of the bounding box provided by DroneVG-R1, thereby serving as a proxy metric for its implicit visual grounding reasoning capability. While bounding box IoU is the direct metric for DVGBench's core grounding task, fine-grained segmentation represents a more advanced understanding of an object's spatial extent. By demonstrating that our model's bounding boxes can effectively bootstrap SAM2, we aim to show that DroneVG-R1 provides high-quality spatial information that supports pixel-level delineation, indicating its potential for facilitating finer-grained spatial understanding beyond coarse localization. This exploration aligns with the long-term goal of advancing from coarse to fine spatial reasoning in VG.

\textbf{Visualization.} 
Figure~\ref{fig:viz_seg}  provides a qualitative comparison of pixel-level predictions generated by our DroneVG-R1-7B model and several competing methods on a diverse set of challenging implicit queries from the DVGBench dataset. The GT masks are annotated in red, while the model predictions are overlaid in pink. The visual results compellingly demonstrate the superior performance of our proposed method. In contrast, all other baseline models exhibit significant limitations. Specialized RS models struggle with tasks requiring high-level semantic reasoning, often failing to disambiguate the primary subject of the query. General-purpose segmentation models frequently generate incorrect or entirely missing masks, highlighting their inability to comprehend implicit intent and complex scene dynamics.

\textbf{Results on DVGBench.} Table~\ref{tab:res_overall} shows that DroneVG-R1 exhibits superior performance in the pixel-level VG task. The proposed DroneVG-R1 7B model demonstrates a commanding lead, establishing a new state-of-the-art. It achieves remarkable scores of 46.70\% mIoU, 44.15\% oIoU, and 51.78\% Acc@0.5. This performance constitutes a significant advancement, outperforming the strongest comparative model, GeoPixel, by a large margin of over 17 percentage points in mIoU (29.63\% vs. 46.70\%). The results unequivocally indicate that our method excels in tasks requiring fine-grained spatial understanding and accurate mask generation. An important analysis revolves around the role of reasoning capability, denoted by the Rationale column. While both DroneVG-R1 and other rationale-enabled models (GeoPixel, GeoPix) incorporate reasoning processes, the substantial performance gap highlights that the quality and task-alignment of the rationale are paramount. 

Furthermore, the results reveal that several specialized models tailored for RS (e.g., RMSIN, SegEarth-R1) or segmentation (e.g., LISA, RemoteSAM) struggle on this complex task, with mIoU scores generally below 25\%. This underscores the heightened difficulty of DVGBench's implicit query-based pixel-level VG setting. The fact that a general-purpose LVLM framework, augmented with our method, substantially outperforms these specialized solutions underscores the effectiveness and generality of our approach in integrating semantic reasoning with dense prediction.

In conclusion, the experimental results on pixel-level VG confirm the superior capability of the DroneVG-R1 model. It not only achieves the highest accuracy but also demonstrates that a reasoning-driven approach, when properly optimized, can yield exceptional performance on a task that integrates high-level cognition with low-level pixel acuity.

% \subsection{Visual Analysis}

% Figure~\ref{fig:viz_seg} demonstrates that our method achieves superior segmentation performance compared to existing general-domain or RS domain pixel-level LVLMs. By simply connecting a segmentation model afterward, DroneVG-R1 enables more accurate semantic understanding of referring expressions, indicating that the bottleneck in current pixel-level implicit VG tasks lies in the localization rather than the segmentation. 

\subsection{Ablation Study}

\begin{table}
\caption{Performance (Acc@0.5\%) under different reward functions. This result indicates that the implicit-to-explicit strategy even outperforms direct supervision of bounding boxes.}
\resizebox{\columnwidth}{!}{
  \centering
  \setlength{\tabcolsep}{12pt}
  \begin{tabular}{cccc}
    \toprule
    Format & Perception & Reasoning   & Acc@0.5 \\
    \midrule
    \icoyes & \icono & \icono  &  51.32\%   \\
    \icoyes & \icoyes & \icono  & 53.04\%   \\
    \rowcolor{myblue}\icoyes &  \icono  & \icoyes & 54.07\%   \\
    \icoyes & \icoyes   & \icoyes & \textbf{54.75\%}   \\
    \bottomrule
  \end{tabular}}
    \label{tab:ae_1}
    \vspace{-0.3cm}
\end{table}

\textbf{Effect of Reward Functions.} 
An ablation study was conducted to dissect the contribution of each component within the reward function of the proposed I2E-CoT framework, with the results summarized in Table~\ref{tab:ae_1}. The performance is evaluated using the overall Acc@0.5metricon the region-level VG task. The results clearly demonstrate that each component — Format, Perception, and Reasoning — contributes positively to the final model performance. A particularly noteworthy finding is the substantial individual contribution of the Reasoning reward. The configuration combining Format and Reasoning rewards achieves an accuracy of 54.07\%, which is not only higher than the baseline but also surpasses the configuration that includes Perception but excludes Reasoning. This indicates that the reward signal for generating a correct explicit referrence is, in itself, a powerful driver of accurate grounding, potentially even more critical than the perception reward in certain contexts.

The results robustly validate the core premise of the implicit-to-explicit strategy employed in our method.  This confirms that guiding the model through a reasoning process to arrive at the bounding box is fundamentally more effective than directly supervising the box coordinates.

 % A key finding merits emphasis: under the baseline condition of a default format reward, enhanced efficacy was observed for the reasoning reward function relative to the perception reward function. 

\begin{figure}
    \centering
    \includegraphics[width=0.48\textwidth]{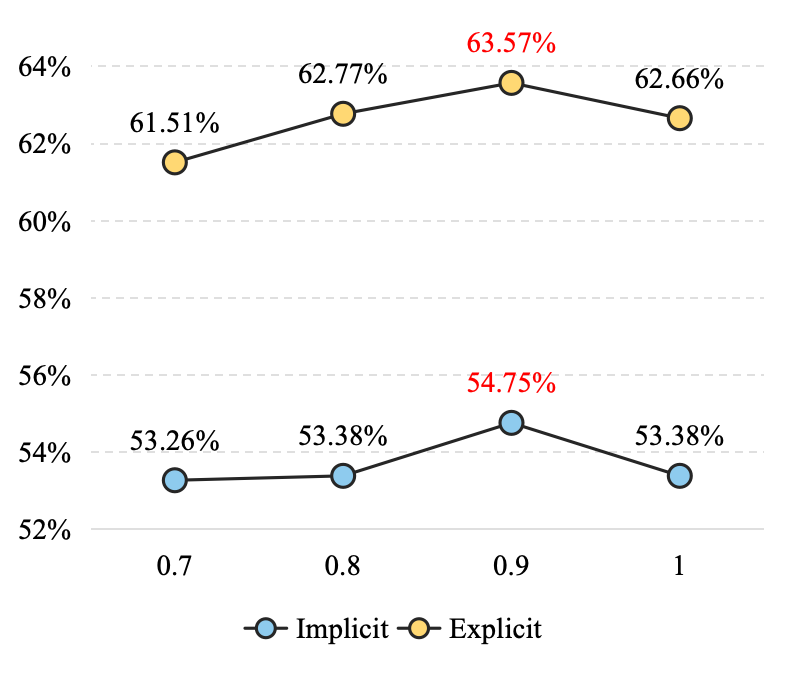}
    \caption{Effect of explicit text similarity threshold on DroneVG-R1's region-level performance (Acc@0.5\%).  
    }
    \label{fig:i2e-thr}
    \vspace{-0.3cm}
\end{figure}

\textbf{Effect of Text Similarity Threshold.} 
Figure~\ref{fig:i2e-thr} illustrates the impact of the explicit text similarity threshold on the region-level grounding accuracy (Acc@0.5\%) of the DroneVG-R1 model, evaluated separately on explicit and implicit queries. A key finding is that the optimal performance for both query types is achieved at an intermediate threshold of 0.9, rather than at the maximum value of 1.0. For explicit queries, the accuracy peaks at 63.57\% with a threshold of 0.9, subsequently declining to 62.77\% at a threshold of 0.8. A similar trend is observed for the more challenging implicit queries, where the maximum accuracy of 54.75\% is attained at the 0.9 threshold. This indicates that insisting on a perfect textual match between the generated rationale and a reference answer is suboptimal. A threshold of 0.9 provides a beneficial tolerance for semantically correct but lexically diverse reasoning patterns, thereby encouraging more robust generalization. The fact that the same threshold value (0.9) is optimal for both query types is of practical significance, as it simplifies parameter tuning by suggesting a universal setting for the similarity threshold regardless of query complexity.

\begin{figure}
 \centering
 % \vspace{-1.mm}
\includegraphics[width=1.0\linewidth]{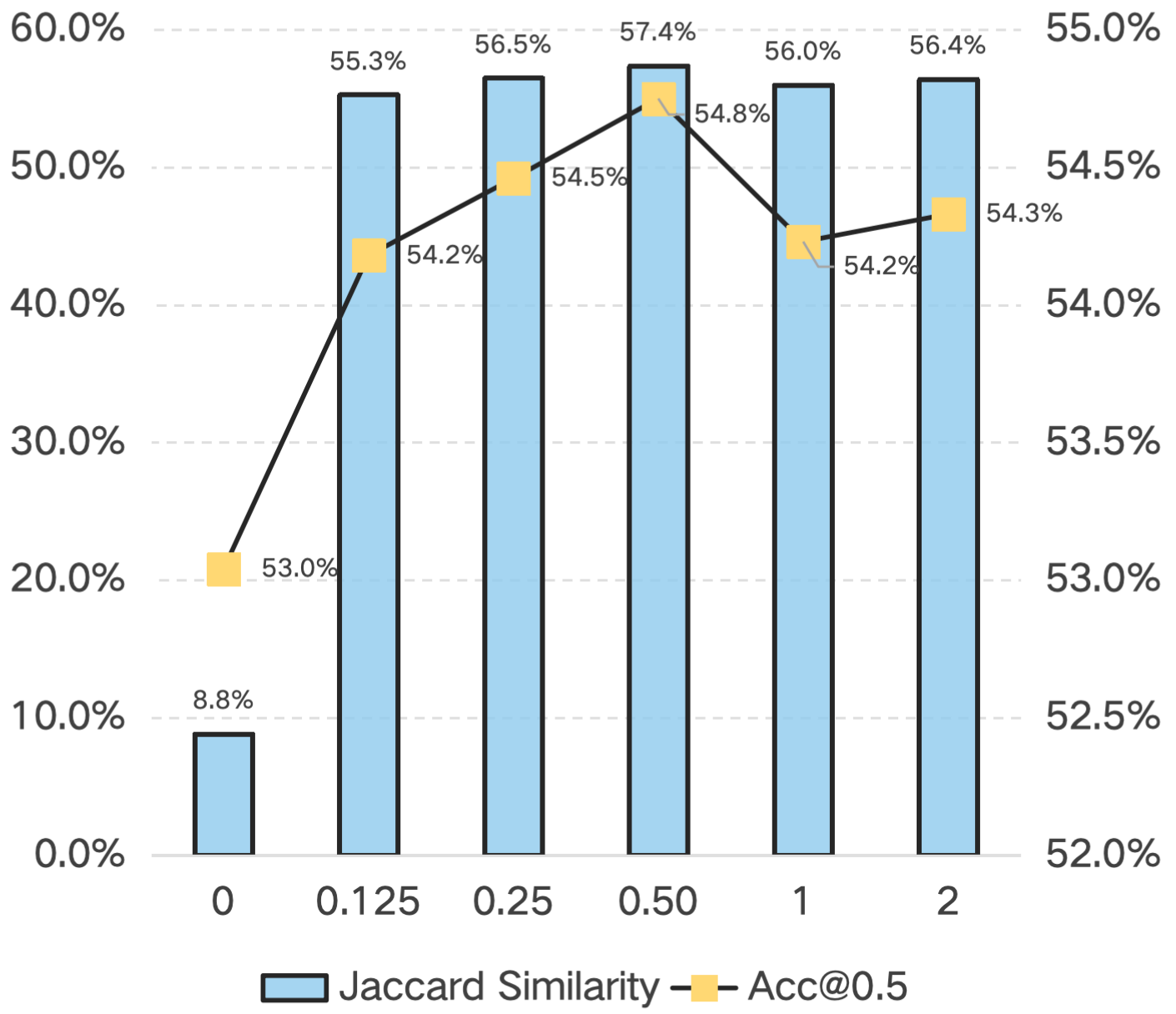}
% \vspace{-2mm}
 \caption{Sensitivity analysis of weight coefficients for reasoning reward.}
 \label{fig:weight_ratio}
 \vspace{-0.3cm}
\end{figure}

\textbf{Effect analysis of weight coefficients for reasoning reward.} A sensitivity analysis in Figure~\ref{fig:weight_ratio} is conducted specifically on the weighting coefficient assigned to the reasoning reward. The Jaccard similarity metric is employed to measure the textual similarity between the explicit references obtained from the model's I2E-CoT process and the human-annotated explicit references. The Acc@0.5 metric gauges the model's performance on the DVGBench. The results reveal that introducing the reasoning reward with any positive weight significantly influences Jaccard similarity, compared to scenarios where it is omitted. Conversely, the final accuracy remains largely unaffected by the specific value of this weight once the reasoning reward is incorporated. In addition, we observed a positive correlation between text similarity and model accuracy, where higher text similarity corresponds to higher model accuracy. This further validates the effectiveness of the I2E-CoT. However, as the weight of the reasoning reward increased, the similarity did not show a monotonic rise. The similarity reached its maximum value at a weight of 0.5. In our future work, we may explore how to overcome the limitations of current explicit text similarity measures by upgrading reinforcement learning algorithms. Based on the above analysis, a default weight of 0.5 was selected for the reasoning reward.

\begin{figure}
    \centering
    \includegraphics[width=0.48\textwidth]{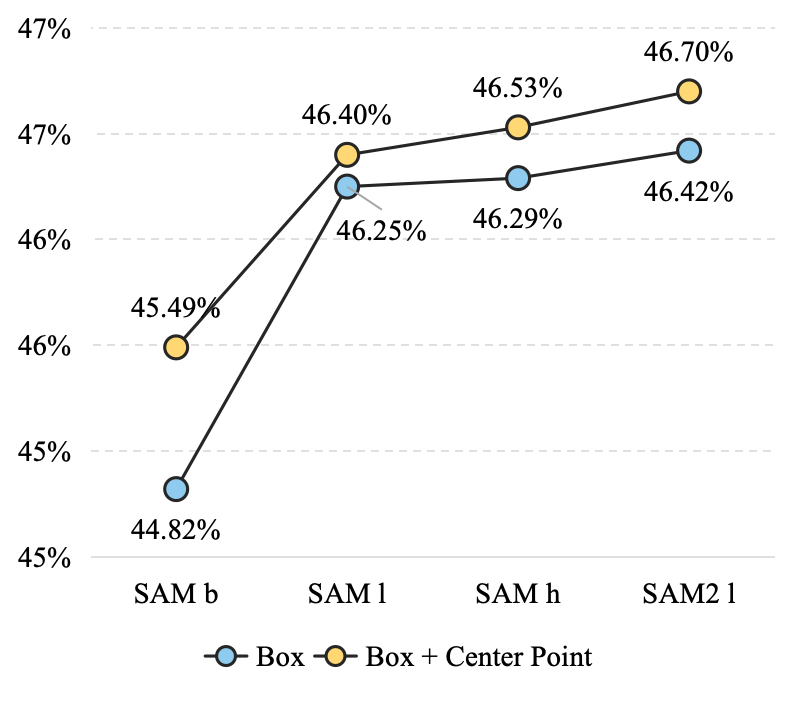}
    \caption{Effect of the segmentation models and prompt methods on DroneVG-R1's pixel-level performance (mIoU).
    }
    \label{fig:sam-ab}
    \vspace{-0.2cm}
\end{figure}

\begin{figure}
    \centering
    \includegraphics[width=0.49\textwidth]{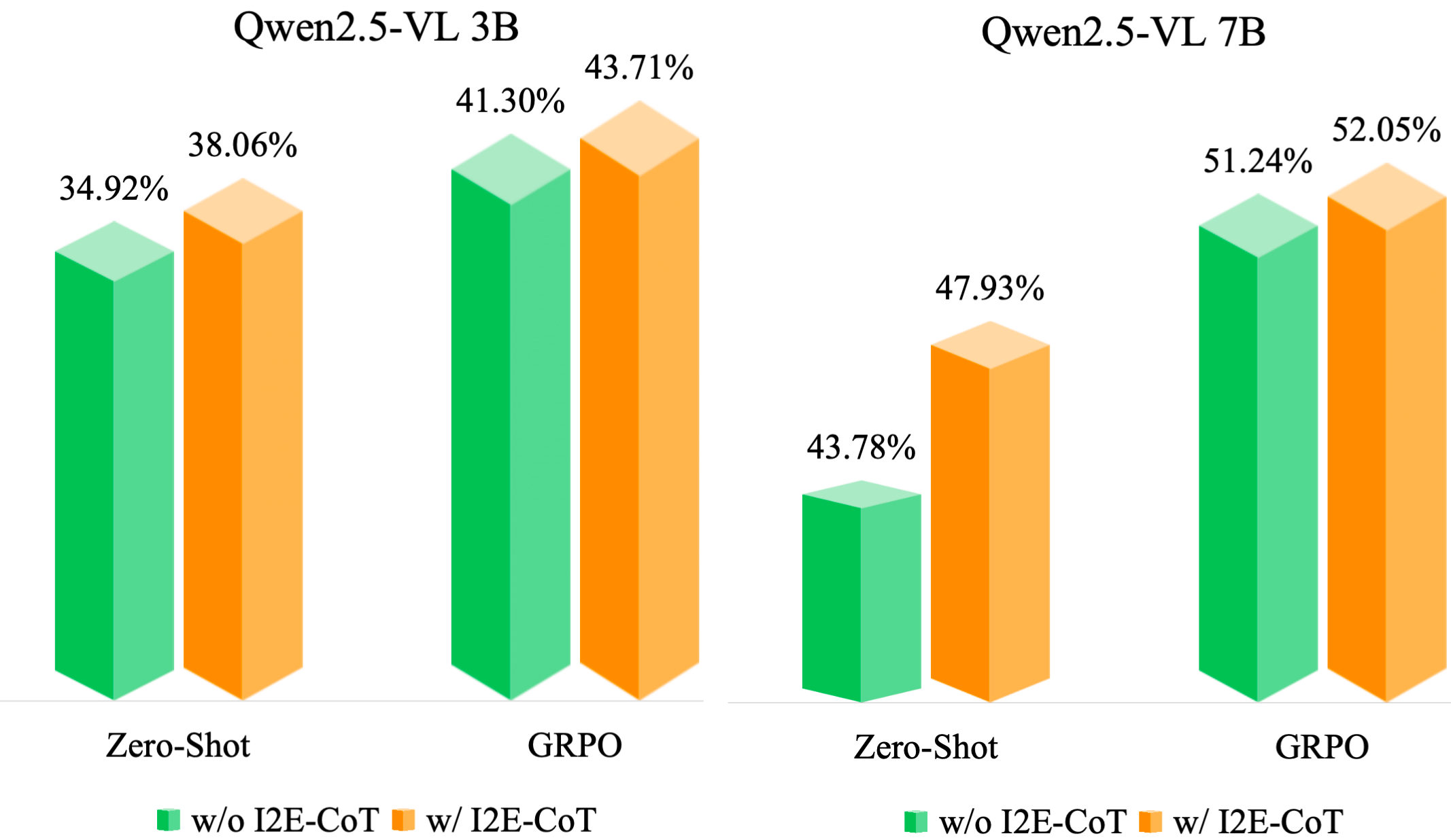}
    \caption{Performance (Acc@0.5\%) under zero-shot and GRPO settings. I2E-CoT enhances both GRPO reinforcement fine-tuning and zero-shot performance. 
    }
    \label{fig:zs}
    \vspace{-0.3cm}
\end{figure}

\textbf{Effect of Segmentation Models.} 
Figure~\ref{fig:sam-ab} illustrates the impact of different segmentation models and prompt methods on the pixel-level performance of the DroneVG-R1, measured by the mIoU metric. The segmentation models evaluated include SAM b, SAM l, SAM h, and SAM2 l, while the prompt methods consist of a bounding box alone and a bounding box augmented with a center point. A key finding is that the "Box + Center Point" prompting strategy consistently yields superior performance compared to the "Box" method across all models. The highest mIoU of 46.70\% is achieved by combining the SAM2 l model with the "Box + Center Point" prompts.  The results demonstrate that providing an additional center point offers more precise spatial guidance to the segmentation model, effectively reducing localization ambiguity within the proposed region and leading to a more accurate segmentation mask.

\begin{figure*}
\
\fontsize{9.0pt}{\baselineskip}\selectfont
\linespread{0.9}\selectfont
%%%%%%%%%%%%%%%%%%%%%%%%%%%%%%%%%%%%%%%%%%%%%%%%%%%%%%%%%
\begin{mybody}
\begin{minipage}[s][5.0cm]{0.28\columnwidth}
\includegraphics[height=5.cm]{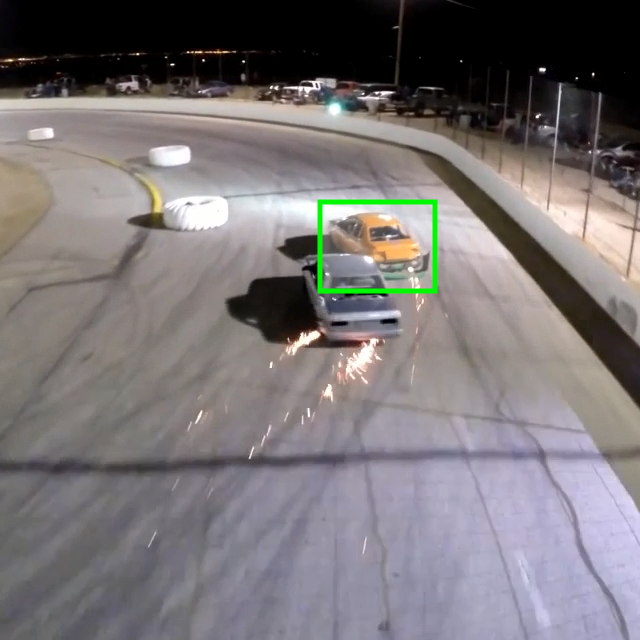}
\end{minipage}
\hspace{1mm}
\begin{minipage}[s][4.6cm]{0.61\columnwidth}
\textbf{Question ID:} \\
460 \\
\textbf{Class:} \\
Sport \\
\textbf{Implicit Query:} \\
The vehicle that was rear-ended \\
\textbf{Ground truth:}\\
\textbf{[327, 212, 424, 282]}
\end{minipage}
\end{mybody}
%%%%%%%%%%%%%%%%%%%%%%%%%%%%%%%%%%%%%%%%%%%%%%%%%%%%%%%%%
\begin{mybody}
\textbf{DeepSeek-VL2-Tiny} \\
\textbf{Response of Implicit Query:}
<|ref|>The vehicle that was rear-ended<|/ref|><|det|>\red{[[400, 380, 632, 572]]}<|/det|>

\end{mybody}
%%%%%%%%%%%%%%%%%%%%%%%%%%%%%%%%%%%%%%%%%%%%%%%%%%%%%%%%%
\begin{mybody}
\textbf{InternVL3.5-8B}\\
\textbf{Response of Implicit Query:}
<think>The image shows two cars on a track, with one car producing sparks as it slides. The car in front, which is yellow, appears to be the one that the other car collided with. The silver car is trailing and seems to have been rear-ended, as indicated by the sparks from its rear tires, suggesting a loss of control.So, the vehicle that was rear-ended is the silver car.</think><answer>\red{[[465, 395, 600, 533]]}</answer>

\end{mybody}
%%%%%%%%%%%%%%%%%%%%%%%%%%%%%%%%%%%%%%%%%%%%%%%%%%%%%%%%%
\begin{mybody}
\textbf{Qwen2.5-VL-7B}\\
\textbf{Response of Implicit Query:}
<think>To determine which vehicle was rear-ended, we need to identify the vehicle that appears to be in motion and has visible sparks or debris trailing behind it, indicating a recent collision.1. The orange car is positioned slightly ahead of the silver car.2. The silver car has visible sparks trailing from its rear, suggesting it may have been rear-ended by the orange car.3. The orange car does not appear to have any visible signs of damage or debris trailing from its rear. Based on these observations, the silver car is the one that was likely rear-ended.</think><answer>\red{[301,254,406,347]}</answer>

\end{mybody}

%%%%%%%%%%%%%%%%%%%%%%%%%%%%%%%%%%%%%%%%%%%%%%%%%%%%%%%%%
\begin{mybody}
\textbf{DroneVG-R1-7B}\\
\textbf{Response of Implicit Query:}
<think> The vehicle that was rear-ended is the one with the orange hood.</think><explicit> orange hood</explicit><answer>\green{[329,210,435,282]}</answer>

\end{mybody}
\caption{The responses of different models to an implicit query in DVGBench.}
\label{fig:example_implicit}
\end{figure*}

In summary, the results confirm that the pixel-level performance of the DroneVG-R1 framework is dependent on two major factors: the quality of the prompts generated by the reasoning model and the capability of the foundational segmentation model. As shown in Table~\ref{tab:res_overall}, the combination of a high-quality segmentation model like SAM2 l with detailed prompt information proves to be the most effective configuration for achieving optimal grounding accuracy. This analysis provides clear guidance for system optimization, emphasizing the need for both precise region prompting and a powerful segmentation backbone.

\subsection{Analysis}

% \subsubsection{Zero-Shot v.s. GRPO}

% \textbf{I2E-CoT serves not only as a GRPO enhancer but also as a training-free technology.}
\textbf{I2E-CoT plays multiple roles:}
The effectiveness of the proposed I2E-CoT strategy is systematically evaluated under both zero-shot and GRPO fine-tuned settings, with the results illustrated in Figure~\ref{fig:zs}. The performance, measured by Acc@0.5\%, is reported for Qwen2.5-VL models of two scales. In the zero-shot setting, applying I2E-CoT boosts the accuracy of the Qwen2.5-VL 3B model from 34.92\% to 38.06\%, and the 7B model from 43.78\% to 47.93\%. This demonstrates that I2E-CoT serves as a powerful and training-free enhancement, effectively eliciting a model's inherent but underutilized reasoning capabilities for complex implicit VG tasks simply by restructuring the inference-time prompt. Furthermore, the I2E-CoT strategy exhibits a strong synergistic effect with models fine-tuned using GRPO framework. For the GRPO-optimized Qwen2.5-VL 3B model, I2E-CoT further increases the accuracy from 41.30\% to 43.71\%. Similarly, for the 7B model, the accuracy is elevated from 51.24\% to a peak performance of 52.05\%. This indicates that I2E-CoT is not rendered redundant by GRPO fine-tuning; instead, it provides an optimal inference-time pathway that allows the fine-tuned model to fully leverage its optimized reasoning capacity, thereby achieving the highest possible performance. In conclusion, the experimental results validate I2E-CoT as a highly effective and versatile strategy. It significantly enhances performance both as a zero-shot tool for unleashing latent model capabilities and as a complementary inference method that pushes the limits of models already advanced by reinforcement fine-tuning.

\textbf{I2E-CoT reduces the difficulty of implicit VG:}
Figure~\ref{fig:example_implicit} analysis examines the performance of four contemporary LVLMs — DeepSeek-VL2-Tiny, InternVL3.5-8B, Qwen2.5-VL-7B, and DroneVG-R1-7B — on a challenging implicit query from the DVGBench dataset. The task requires identifying the bounding box coordinates of "the vehicle that was rear-ended" based on a dynamic racing scene. The results reveal significant disparities in the models' abilities to perform simultaneous visual perception, spatial reasoning, and textual justification. For DeepSeek-VL2-Tiny, it loses its reasoning ability on the VG task and can only simply repeat the referring text in the question, resulting in incorrect answers. In contrast, InternVL3.5-8B achieves a fine-grained understanding of image details, noticing sparks at the rear of the silver vehicle and thus inferring that the silver vehicle is out of control. However, the loss of control of the silver vehicle was not caused by being rear-ended; rather, it lost control first, which led to it rear-ending the yellow vehicle, making the yellow vehicle the victim. Qwen2.5-VL-7B provides a more detailed analysis of the image content, but both it and InternVL3.5-8B misunderstand the logical relationship between rear-ending and being rear-ended. Although the proposed Drone-VG-R1-7B model provides a relatively concise answer, it avoids the need for complex analysis of the rear-end collision by converting implicit references into explicit "orange hood," significantly reducing the difficulty of the task. We believe that the reasoning process serves the model, and a reasoning process suitable for the model is not necessarily one that is human-preferred.

\begin{figure*}
    \centering
    \includegraphics[width=1.0\textwidth]{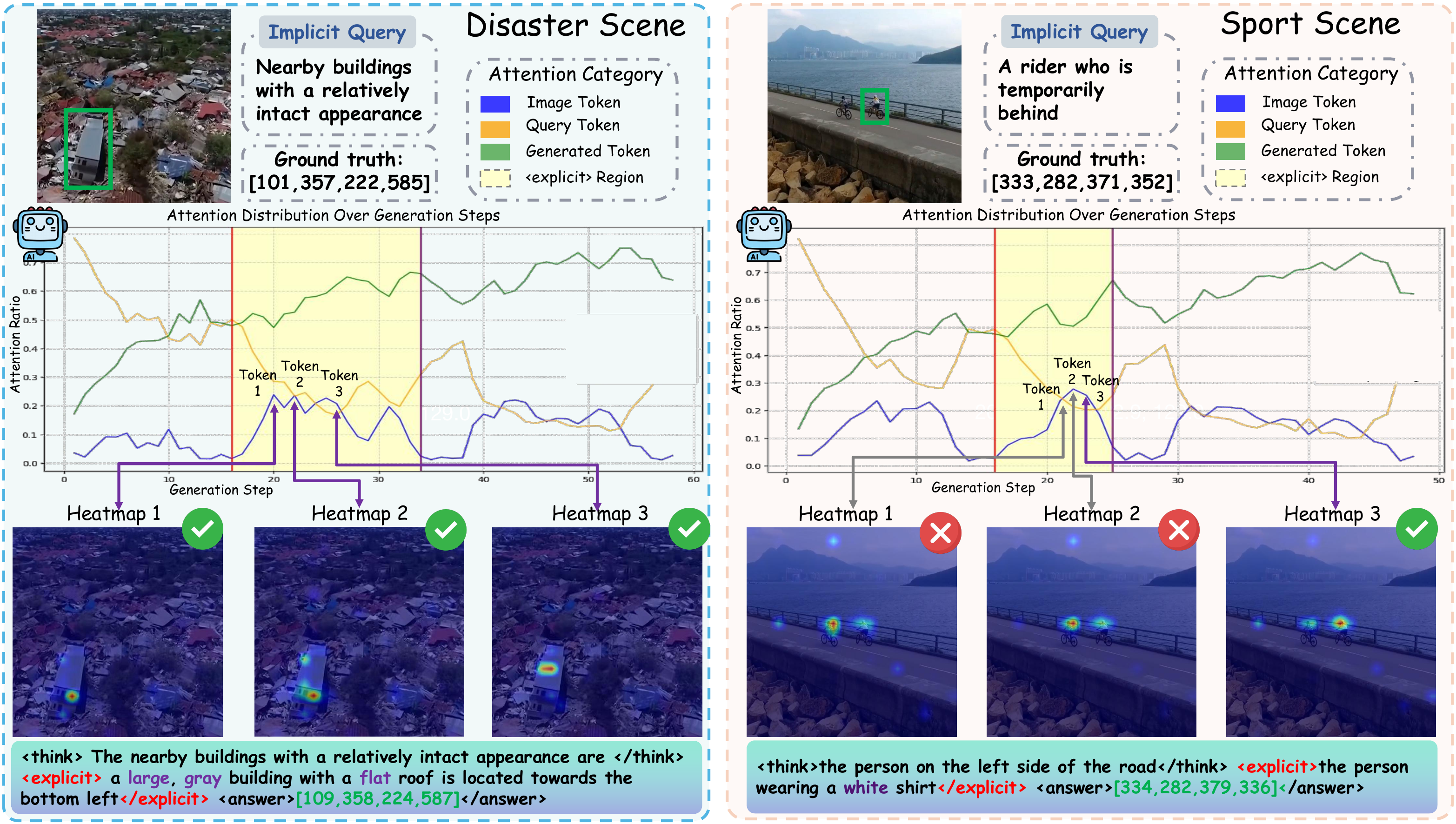}
    \caption{Two examples used to demonstrate the impact of the I2E-CoT mechanism on image attention.
    }
    \label{fig:heatmap_demo}
    % \vspace{-0.5cm}
\end{figure*}

% \textbf{I2E-CoT influences the grounding results by altering the model's attention.}
\textbf{Visualization of I2E-CoT's attention heatmaps:} The generation of attention maps is designed to visualize the model's focus on the input image when producing each output token, providing crucial insights into the model's decision-making process. Specifically, for every generated token, we record the attention weights from all layers and all attention heads of the model. These weights, which represent the importance assigned to different parts of the input (including image tokens, question text tokens, and previously generated tokens), are then averaged across all layers and heads to create a single, consolidated attention vector. Subsequently, the portion of this averaged attention vector corresponding specifically to the input image tokens is extracted. This 1D vector, which reflects the aggregate attention paid to the image patches, is then reshaped into a 2D grid that corresponds to the original spatial layout of the image patches, forming a low-resolution attention map. Finally, this 2D map is normalized and resized (upscaled via interpolation) to match the dimensions of the original input image. The resulting heatmap can be directly overlaid onto the image, highlighting the regions that most influenced the generation of the current token.

 As illustrated in Figure~\ref{fig:heatmap_demo}, we plotted the image attention proportion curves and corresponding heatmaps for implicit references from two distinct scenarios to visually demonstrate the impact of the I2E-CoT mechanism. In the disaster scenario example on the left, the implicit query targets nearby buildings with a relatively intact appearance, and the explicit references describe the building’s size, color, and shape features like “large, gray building with a flat roof”. From the attention proportion curves, these explicit feature - related tokens correspond to higher proportions of image attention. By visualizing the heatmaps associated with these words, we found that the model accurately localized the target. Specifically, the heatmap clearly focuses on the intended building, and the model demonstrates an understanding of explicit features such as "flat". These words indeed helped the model precisely locate the target region. This also indicates that generating explicit outputs genuinely alters the model’s image attention rather than being merely a formality. 

 Most strikingly, in the sports scenario example on the right, the implicit query is about a rider who is temporarily behind, and the explicit reference is “the person wearing a white shirt”. We observed a moment of shift in image attention. Before the phrase "white shirt" appeared, the model's attention was somewhat dispersed or focused on the left person. After the phrase "white shirt" appeared, as shown in the attention proportion curves and heatmaps, the model’s image attention shifted significantly from the left person to the rider on the right wearing a white shirt. It was this single explicit descriptive word that refined the model’s localization, making the attention accurately lock onto the target rider. Thus, we posit that I2E-CoT functions by altering the model's attention to the image, guiding it to focus on relevant regions through explicit descriptions.

\textbf{Evaluation of Generalization Capability:} Evaluation on additional benchmarks is essential to demonstrate generalization. To address this, we have supplemented our evaluation with experiments on two public VG datasets in RS: EarthReason \citep{li2025segearth} and GeoPix\citep{ou2025geopix}. EarthReason is a pixel-level RS
implicit VG dataset. For each object, it provides five implicit referring queries, but the specific referring text selected for the evaluation in the paper is not indicated. Therefore, a direct performance comparison between the method proposed in that article and our method is not feasible. Furthermore, we validate that our proposed method achieves performance gains even on out-of-domain data. The EarthReason dataset is considered out-of-domain because it is based entirely on satellite imagery, in contrast to our training samples, which consist exclusively of drone-captured imagery.

\begin{figure*}
    \centering
    \includegraphics[width=1.\textwidth]{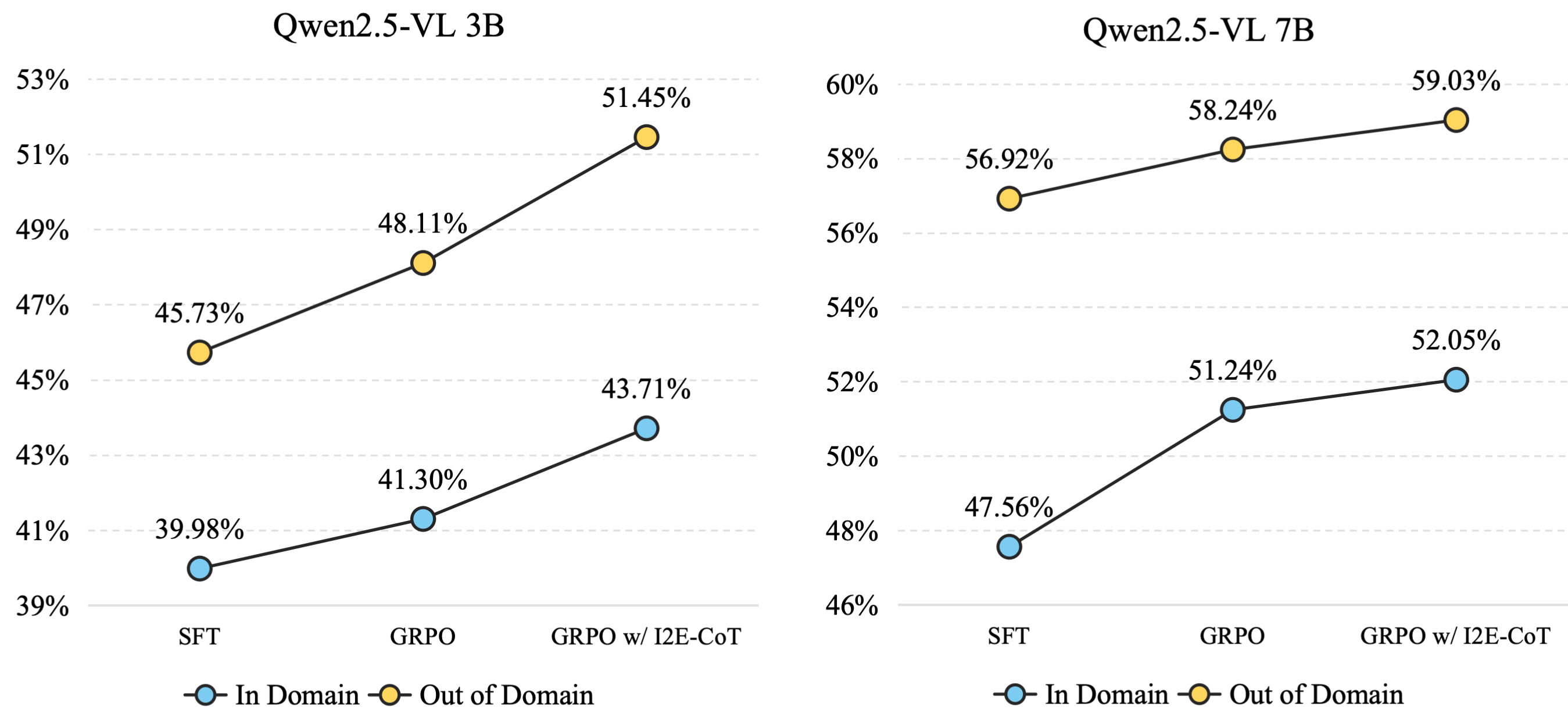}
    \caption{Performance (Acc@0.5\%) under in-domain and out-of-domain settings. With limited training samples, GRPO with I2E-CoT demonstrates better generalization capability compared to SFT and GRPO.}
    \label{fig:domain-ab}
    \vspace{-0.5cm}
\end{figure*}

\begin{table}
  \centering
    \caption{Performance comparison of pixel-level VG on GeoPix datasets.}
\resizebox{1.\columnwidth}{!}{
  \setlength{\tabcolsep}{2pt}
  \begin{tabular}{lcccccc}
    \toprule
    \multirow{2}{*}{Method} & \multicolumn{2}{c}{SIOR-T} &  \multicolumn{2}{c}{FAST-T}
    &  \multicolumn{2}{c}{SOTA-T}\\
    \cline{2-7}
     & mIoU & Acc@0.5 & mIoU & Acc@0.5  & mIoU & Acc@0.5\\
    \midrule
    GeoPix & \textbf{78.97\%} & \textbf{93.71\%}	& 34.99\% &	33.84\%	&27.60\%  &	23.52\%\\  
    \midrule
    Qwen2.5-VL 3B &38.98\%	&39.54\%	&31.18\%	&32.73\%	&21.45\%	&21.27\% \\
    \rowcolor{myblue}DroneVG-R1 3B & 40.88\%	&40.88\%	&35.67\%	&37.85\%	&25.14\%	&25.39\%\\
    Qwen2.5-VL 7B & 46.85\%	&47.89\%	&39.79\%	&42.23\%	&29.49\%	&30.32\% \\
    \rowcolor{myblue}DroneVG-R1 7B & 48.18\%	&50.35\%	&\textbf{41.88\%}	&\textbf{45.06\%}	&\textbf{31.16\%}	&\textbf{32.06\%}\\

    \bottomrule
  \end{tabular}}
  \vspace{-0.3cm}
  \label{tab:geopix}
\end{table}

The generalization capability of the proposed method is rigorously evaluated under both in-domain and out-of-domain settings, with the results for Qwen2.5-VL models of 3B and 7B scales depicted in Figure~\ref{fig:domain-ab}. In-Domain refers to the DVGBench test set proposed in this paper, whereas Out-of-Domain refers to the EarthReason dataset, a reasoning segmentation benchmark that comprises satellite imagery. The corresponding bounding box annotations are derived directly from the segmentation masks. The performance is measured by overall Acc@0.5\% and compares SFT against our proposed GRPO framework, both with and without the I2E-CoT inference strategy. It is worth mentioning that due to the extreme challenge of manually annotating our benchmark, the model's in-domain score is even lower than its out-of-domain score, and we primarily focus on the performance differences among the three methods. GRPO w/ I2E-CoT yields the highest accuracies in both settings, pushing the 7B model's out-of-domain performance to 59.03\%. The experimental results on domain shift validate that the GRPO framework, especially when combined with I2E-CoT inference, effectively mitigates overfitting and promotes the learning of robust and generalizable VG skills.

Since the official repository of GeoPix only provides inference scripts but not evaluation scripts, to ensure fairness, we uniformly used our own evaluation script to test both DroneVG-R1 and GeoPix on the GeoPix validation set, as shown in Table \ref{tab:geopix}. Although the GeoPix paper reports performance on three datasets—SIOR-T, FAST-T, and SOTA-T—these results were obtained using three separate sets of model weights. To date, the authors have only open-sourced the model fine-tuned on SIOR-T (GeoPix-ft-sior\_rsicap). Its superior performance on this dataset is therefore expected and may be attributed to potential overfitting. Given that our main objective is to compare the generalization ability of different models, we focus primarily on GeoPix’s scores on the FAST-T and SOTA-T datasets. As shown in Table \ref{tab:geopix}, the DroneVG-R1 3B model outperforms GeoPix on FAST-T, while the DroneVG-R1 7B
model demonstrates a substantial advantage on both FAST-T and SOTA-T.

\begin{figure*}

% \fontsize{9.0pt}{\baselineskip}\selectfont
\linespread{0.9}\selectfont
%%%%%%%%%%%%%%%%%%%%%%%%%%%%%%%%%%%%%%%%%%%%%%%%%%%%%%%%%
\begin{mybody}
% image: 852
\begin{minipage}[s][5.0cm]{0.51\columnwidth}
\includegraphics[height=5.cm]{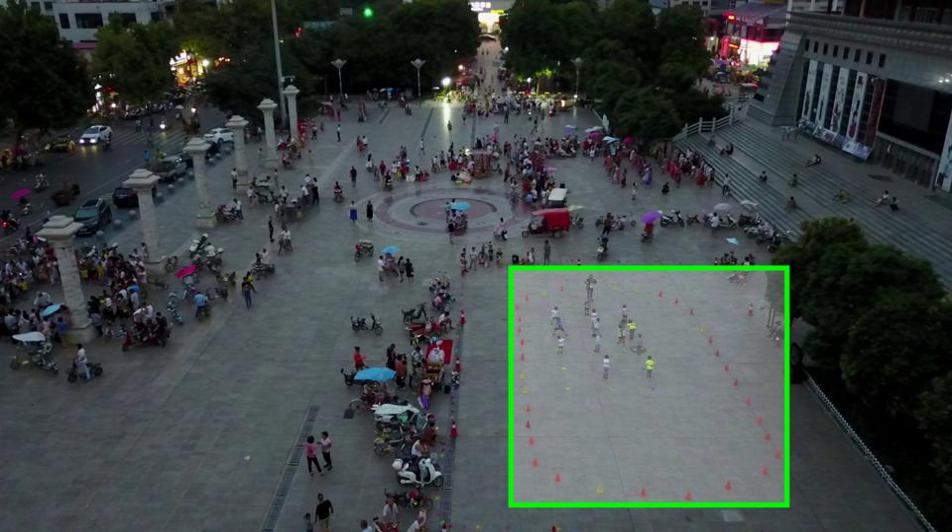}
\end{minipage}
\hspace{1mm}
\begin{minipage}[s][4.6cm]{0.44\columnwidth}
\textbf{Question ID:} 
33 \\
\textbf{Class:} 
Social Activity\\
\textbf{Explicit Query:} \\
The area to the right in the middle, surrounded by barricades. \\
\textbf{Ground truth:}
\textbf{[519,280,778,505]}
\end{minipage}
\end{mybody}
%%%%%%%%%%%%%%%%%%%%%%%%%%%%%%%%%%%%%%%%%%%%%%%%%%%%%%%%%
\begin{mybody}
\textbf{DeepSeek-VL2-Tiny} \\
\textbf{Response of Explicit Query:}
<|ref|>The area to the right in the middle, surrounded by barricades<|/ref|>\red{<|det|>[[0, 0, 999, 999]]<|/det|>}
\end{mybody}
%%%%%%%%%%%%%%%%%%%%%%%%%%%%%%%%%%%%%%%%%%%%%%%%%%%%%%%%%
\begin{mybody}
\textbf{InternVL3.5-8B}\\
\textbf{Response of Explicit Query:}
<think> So, I need to find the area to the right in the middle of the image that's surrounded by barricades. Let me look at the image again. The area in question is probably the section with the red and yellow cones forming a path. The barricades might be the cones themselves, or perhaps some barriers around a specific section.  Looking at the middle-right part of the image, there's a section with a lot of people and some structures. The cones are arranged in a line, creating a path or boundary. The area surrounded by these cones might be what's needed. The coordinates should cover this section.  I'll estimate the coordinates based on the image. The top-left corner of this area might be around the middle of the image, and the bottom-right would extend to the right side where the cones end. Let's say the top-left is approximately at the middle height, and the width goes to the right.  So, the coordinates might be something like [x1, y1, x2, y2] where x1 is the left side of the area, y1 is the top, x2 is the right side, and y2 is the bottom. Let's approximate:  - x1: around 1/3 from the left (since the cones start a bit to the right of center) - y1: middle of the image height - x2: near the right edge - y2: bottom of the image or the lower part where the cones end  But to be precise, I'll look at the image again. The cones start roughly around the middle and go to the right. The top of the area might be around the middle height, and the bottom might be lower, where the cones end. Let's say: Top-left (x1, y1): around (400, 200) Bottom-right (x2, y2): around (800, 600)  These are rough estimates. The exact numbers might vary, but this should cover the area with the cones.</think>\red{<answer>[400, 200, 800, 600]</answer>}
\end{mybody}
%%%%%%%%%%%%%%%%%%%%%%%%%%%%%%%%%%%%%%%%%%%%%%%%%%%%%%%%%
\begin{mybody}
\textbf{Qwen2.5-VL-7B}\\
\textbf{Response of Explicit Query:}
<think>To find "The area to the right in the middle, surrounded by barricades," I need to identify the central part of the image where there is a clear division marked by what appears to be orange cones or similar markers. This area seems to be a designated space, possibly for an event or activity, as it's separated from the rest of the square. The area in question would be the one that is clearly defined by these markers and is located towards the right side of the image.</think>\green{<answer>[476,258,803,531]</answer>}
\end{mybody}

%%%%%%%%%%%%%%%%%%%%%%%%%%%%%%%%%%%%%%%%%%%%%%%%%%%%%%%%%
\begin{mybody}
\textbf{DroneVG-R1-7B}\\
\textbf{Response of Explicit Query:}
<think> The area to the right in the middle, surrounded by barricades</think> <explicit> is a large open space with red and yellow cones marking boundaries.</explicit> \green{<answer>[476,280,835,532]</answer>}
\end{mybody}
\caption{The responses of different models to an explicit query in DVGBench.}
\label{fig:example_explicit}
\end{figure*}

\textbf{Effectiveness validation on explicit VG:}
The study finds that the design of I2E-CoT is not only beneficial for Implicit VG tasks but also enhances the model's performance on general explicit VG tasks. As shown in Figure~\ref{fig:example_explicit}, compared to the convoluted reasoning processes of other models, DroneVG-R1 adopts a more direct approach by supplementing additional detailed features to refine the explicit description of the target region. As revealed in Figure~\ref{fig:heatmap_demo}, it is precisely these key explicit words such as colors that play a critical role in the model's localization.

\begin{table*}[!h]
\caption{Performance Comparison (Acc@0.5\%) on DVGBench across Zero-Shot, Few-Shot, and DroneVG-R1 with Implicit Queries.} 
\setlength{\tabcolsep}{8pt}
\resizebox{\textwidth}{!}{
\begin{tabular}{lccccccc}
\toprule
 Model    & Security & Traffic & Social Activity & Disaster & Productive Activity & Sport & AVG \\ 

\midrule
Qwen2.5-VL 3B     & 45.24\%  & 25.69\% & 29.67\%  & 22.64\%   & 51.24\%  & 35.03\% & 34.92\%   \\
Qwen2.5-VL 3B (2-shots)   & 40.48\%  & 23.26\% & 36.26\%  & 22.64\%   & 44.21\%  & 25.48\% & 32.06\%$\downarrow$    \\
Qwen2.5-VL 3B (5-shots)   & 33.33\%  & 14.24\% & 25.27\%  & 24.53\%   & 40.91\%  & 23.57\% & 26.97\%$\downarrow$   \\
\rowcolor{mylightblue}DroneVG-R1 3B   & 50.00\%  & 42.36\% & 41.76\%  & 30.19\%   & 57.85\%  & 40.13\% & 43.71\%$\uparrow$   \\
\midrule
Qwen2.5-VL 7B    & 35.71\%  & 41.32\% & 34.07\%  &  \textbf{47.17\%}   & 63.64\%  & 40.76\% & 43.78\%   \\
Qwen2.5-VL 7B (2-shots)   & 35.71\%  & 27.43\% & 27.47\%  & 39.62\%   & 48.76\%  & 29.94\% & 34.82\%$\downarrow$    \\
Qwen2.5-VL 7B (5-shots)  & 35.71\%  & 18.75\% & 30.77\%  & 32.08\%   & 45.45\%  & 28.03\% & 31.80\%$\downarrow$  \\
\rowcolor{myblue}DroneVG-R1 7B    & \textbf{57.14\%}  & \textbf{52.08\%}  & \textbf{45.05\%}  & 43.40\%  & \textbf{70.66\%}  & \textbf{43.95\%}  & \textbf{52.05\%}$\uparrow$  \\

\bottomrule
\end{tabular}}

\label{tab:few-shot}
\end{table*}

\textbf{Effect of Few-shot Setting: } We also conducted the few-shot experiments, and the results reveal a key finding: few-shot prompting consistently degrades performance on our challenging implicit reasoning benchmark. As shown in the Table 7, for both 3B and 7B models, adding 2-shot and 5-shot examples lowers the average accuracy compared to the zero-shot baseline. This indicates that the complexity and diversity of our implicit questions make them unsuitable for few-shot generalization. In stark contrast, DroneVG-R1 achieve significantly higher accuracy, demonstrating a clear advantage over prompt-based methods. This performance gap strongly substantiates the necessity of our proposed GRPO with I2E-CoT method.

The significant advantage of GRPO over few-shot prompt tuning originates from its fundamental improvements in three key aspects: optimization stability, generalization capability, and reasoning consistency, making it particularly suitable for the complex and diverse implicit reasoning required by our tasks. The GRPO objective function incorporates policy ratio clipping and KL-divergence regularization, which jointly ensure that policy updates do not deviate excessively from the initial policy or undergo drastic fluctuations. This mechanism effectively prevents model collapse or performance degradation during optimization, providing a guarantee for training stability that is absent in few-shot prompting. GRPO directly optimizes the final objectives we care about through its reward function. This goal-driven learning approach enables the model to explore and internalize effective reasoning paths, rather than passively adapting to a limited set of in-context examples. The reward function of GRPO can be explicitly designed to encourage the generation of interpretable reasoning processes that are logically coherent and clearly structured. By computing rewards and performing optimization at the level of the complete output sequence, GRPO encourages the model to produce globally consistent reasoning content, ensuring that each step logically leads to the final answer, rather than focusing solely on local token prediction.

\section{Discussion}

\begin{figure}
 \centering
\includegraphics[width=1.\linewidth]{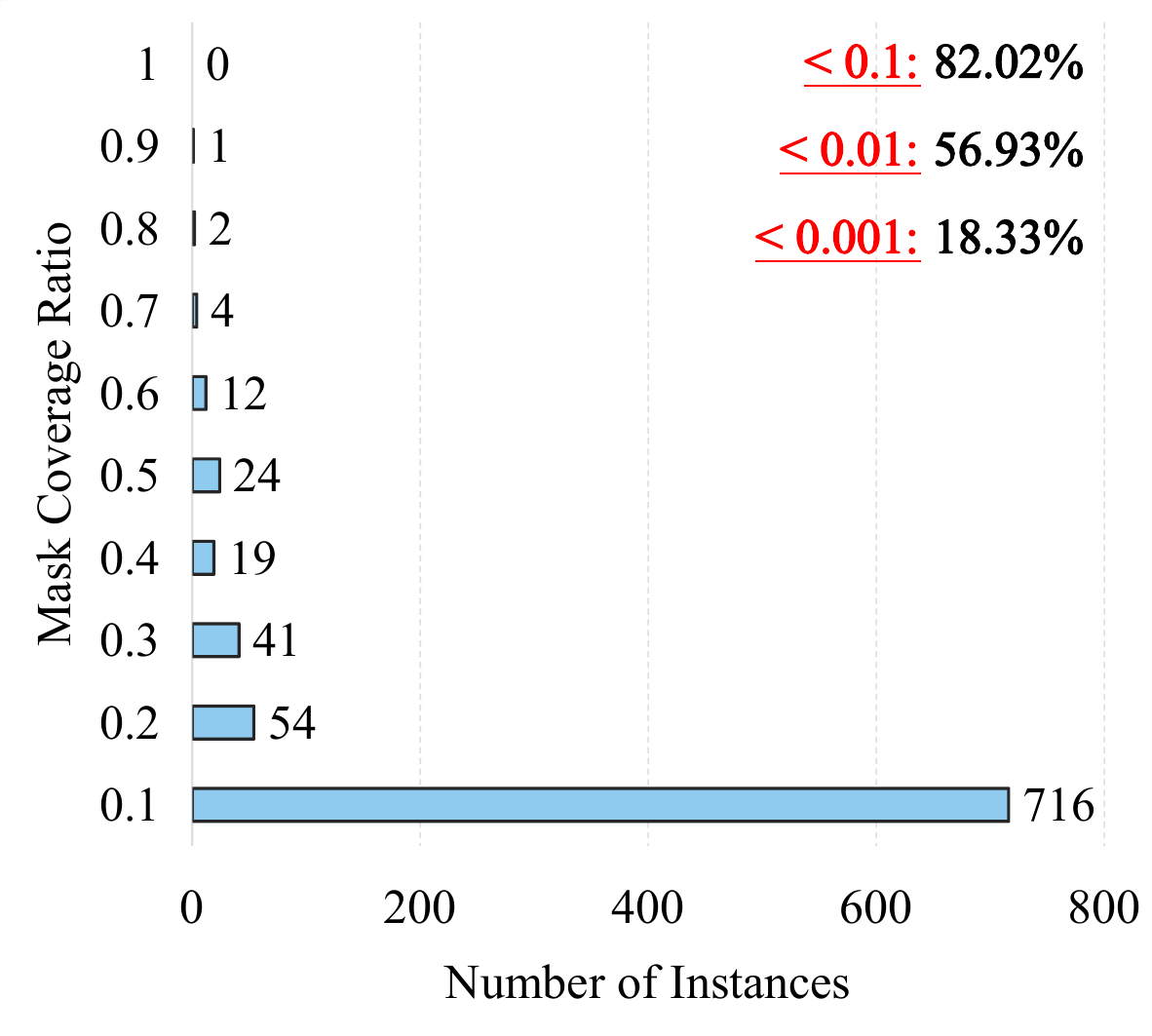}
 \caption{The distribution of instance number by mask coverage ratio in DVGBench. Instances with lower coverage have greater difficulty.}
 \label{fig:mask_ratio}
\end{figure}

\begin{figure}
 \centering
\includegraphics[width=1.0\linewidth]{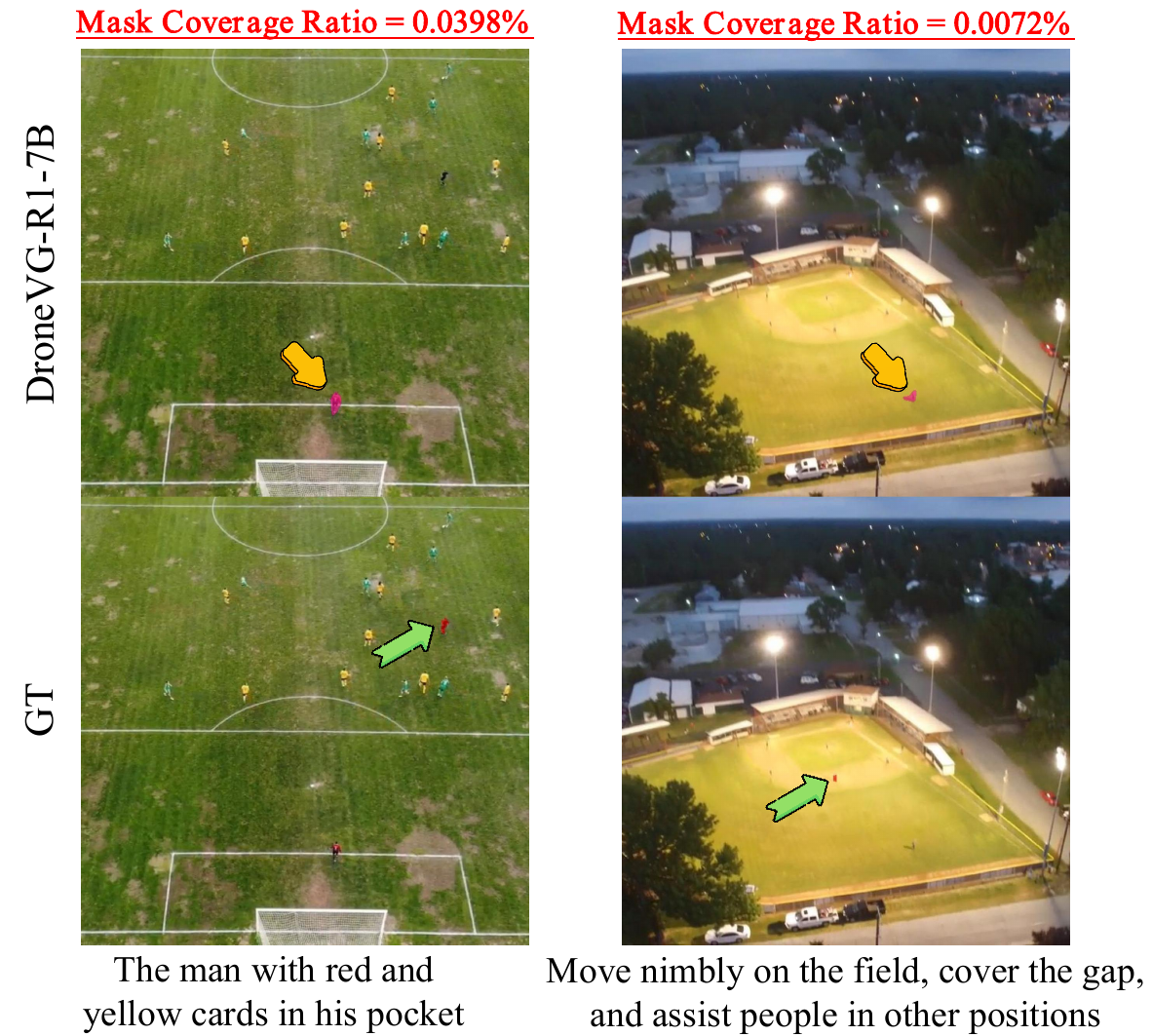}
 \caption{Ultra-small objects pose significant challenges for pixel-level VG task.}
  \label{fig:bad_cases}
\end{figure}

As shown in Figure~\ref{fig:mask_ratio} we analyzed the area ratio of the referred object's mask to the entire image in our dataset. Statistical analysis reveals a notable prevalence of very small instances in the dataset, where the Mask Coverage Ratio falls below 0.1 in 82.02\% of cases, below 0.01 in 56.93\%, and even below 0.001 in 18.33\% of instances. The significant proportion of these minuscule targets poses considerable challenges for visual localization, primarily due to the limited pixel area available for extracting discriminative features, which can adversely affect the precision of both bounding box regression and mask generation. Our models still struggle to accurately identify referred objects that occupy a very small area ratio. Figure~\ref{fig:bad_cases} presents several failure examples, where the combined pressure from the reasoning demands of implicit reference and the localization challenges inherent to small targets leads to model failures. A promising future research direction for addressing the implicit visual grounding of small targets may involve agent frameworks that integrate zoom-in tools.

Furthermore, although we have validated the effectiveness of our method on a manually annotated dataset comprising over 2,863 samples, the dataset size remains limited, which restricts significant performance improvements. Scaling up the collection of high-quality implicit-to-explicit VG data is essential for further enhancing the reasoning capabilities of UAV agents. In the future, self-play reinforcement learning could serve as an effective approach to substantially increase the volume of implicit-to-explicit visual grounding data. This technique involves one model generating questions and another answering them, allowing for mutual evaluation and collaborative advancement.

\section{Conclusion}

In this study, we tackled the critical challenge of implicit reasoning in visual grounding (VG) for UAV remote sensing (RS) imagery. To systematically diagnose and enhance this capability, we introduced \dataset, a novel benchmark comprising human-annotated explicit and implicit queries across diverse aerial scenes. The dual-query design of \dataset provides a principled means to evaluate not just perception, but the reasoning gap in VG models. Driven by the insights from this benchmark, we proposed the Implicit-to-Explicit Chain-of-Thought (I2E-CoT) strategy. This method effectively mitigates the difficulty of implicit VG by decomposing the task into a reasoning step (generating an explicit description) followed by a grounding step. Our analysis, supported by attention visualization, confirms that I2E-CoT successfully guides the model's focus to the relevant regions, leading to more accurate localization. By further integrating a dedicated reasoning reward and leveraging GRPO during training, we developed DroneVG-R1, an LVLM capable of handling both region-level and pixel-level implicit VG tasks. Extensive experimental results demonstrate the superiority of our proposed framework. The comparative analysis also reveals the limitations of existing models, underscoring the necessity of specialized benchmarks and methods for implicit VG in RS. The findings of this study highlight the importance of explicit reasoning pathways for implicit VG tasks in UAV applications. The proposed \dataset, methodology, and the DroneVG-R1 model establish a solid foundation for future research in implicit VG of RS imagery. Future work will focus on expanding the benchmark to include more complex, large-scale scenarios and exploring the integration of advanced reasoning models to achieve a more generalized and robust implicit VG capability.

\appendix

\section{More Details about Annotation Process}

Our annotators began by reviewing the drone images from these two datasets to identify those suitable for creating questions, based on the presence of distinctive referential objects in the images. We then proceeded to develop questions involving implicit references. Then, annotators were instructed to construct referring texts that require domain-specific knowledge to understand, taking into account the context captured in the drone imagery. For traffic scene drone images, we developed questions related to traffic regulations and vehicles. For disaster scenarios, the focus was on hazard zone identification and post-disaster rescue-related references. For sports settings, questions were designed around the rules of the respective sports. In production environments, questions were formulated based on the functional roles of objects within the operational workflow. For public security scenarios, questions were constructed from a law enforcement perspective, targeting potential criminal subjects. For social activity scenes, references involved reasoning about the ongoing event. For other urban drone images, distinctive buildings, vehicles, or individuals were selected as referential objects for question creation. After completing the implicit referring texts, annotators then created explicit referring texts for the same targets, describing them using salient visual features such as color and relative location. It is important to note that no semi-automated assistance methods, such as text overlap matching, were used in this process.

\printcredits

\bibliographystyle{cas-model2-names}

\bibliography{cas-refs}

@inproceedings{sun2022visual,
  title={Visual grounding in remote sensing images},
  author={Sun, Yuxi and Feng, Shanshan and Li, Xutao and Ye, Yunming and Kang, Jian and Huang, Xu},
  booktitle={Proceedings of the 30th ACM International conference on Multimedia},
  pages={404--412},
  year={2022}
}

@article{yuan2024rrsis,
  title={Rrsis: Referring remote sensing image segmentation},
  author={Yuan, Zhenghang and Mou, Lichao and Hua, Yuansheng and Zhu, Xiao Xiang},
  journal={IEEE Transactions on Geoscience and Remote Sensing},
  year={2024},
  publisher={IEEE}
}

@article{zhan2023rsvg,
  title={Rsvg: Exploring data and models for visual grounding on remote sensing data},
  author={Zhan, Yang and Xiong, Zhitong and Yuan, Yuan},
  journal={IEEE Transactions on Geoscience and Remote Sensing},
  volume={61},
  pages={1--13},
  year={2023},
  publisher={IEEE}
}

@article{lan2024lqvg,
  author={Lan, Meng and Rong, Fu and Jiao, Hongzan and Gao, Zhi and Zhang, Lefei},
  journal={IEEE Transactions on Geoscience and Remote Sensing}, 
  title={Language Query-Based Transformer With Multiscale Cross-Modal Alignment for Visual Grounding on Remote Sensing Images}, 
  year={2024},
  volume={62},
  number={},
  pages={1-13}}

@article{li2024lgpa,
  author={Li, Ke and Wang, Di and Xu, Haojie and Zhong, Haodi and Wang, Cong},
  journal={IEEE Transactions on Geoscience and Remote Sensing}, 
  title={Language-Guided Progressive Attention for Visual Grounding in Remote Sensing Images}, 
  year={2024},
  volume={62},
  number={},
  pages={1-13}}

@inproceedings{liu2024rotated,
  title={Rotated multi-scale interaction network for referring remote sensing image segmentation},
  author={Liu, Sihan and Ma, Yiwei and Zhang, Xiaoqing and Wang, Haowei and Ji, Jiayi and Sun, Xiaoshuai and Ji, Rongrong},
  booktitle={Proceedings of the IEEE/CVF Conference on Computer Vision and Pattern Recognition},
  pages={26658--26668},
  year={2024}
}

@inproceedings{kuckreja2024geochat,
  title={Geochat: Grounded large vision-language model for remote sensing},
  author={Kuckreja, Kartik and Danish, Muhammad Sohail and Naseer, Muzammal and Das, Abhijit and Khan, Salman and Khan, Fahad Shahbaz},
  booktitle={Proceedings of the IEEE/CVF Conference on Computer Vision and Pattern Recognition},
  pages={27831--27840},
  year={2024}
}

@article{li2024vrsbench,
  title={Vrsbench: A versatile vision-language benchmark dataset for remote sensing image understanding},
  author={Li, Xiang and Ding, Jian and Elhoseiny, Mohamed},
  journal={Advances in Neural Information Processing Systems},
  volume={37},
  pages={3229--3242},
  year={2024}
}

@article{zhou2024geoground,
  title={Geoground: A unified large vision-language model for remote sensing visual grounding},
  author={Zhou, Yue and Lan, Mengcheng and Li, Xiang and Feng, Litong and Ke, Yiping and Jiang, Xue and Li, Qingyun and Yang, Xue and Zhang, Wayne},
  journal={arXiv preprint arXiv:2411.11904},
  year={2024}
}

@article{yao2025remotesam,
  title={RemoteSAM: Towards Segment Anything for Earth Observation},
  author={Yao, Liang and Liu, Fan and Chen, Delong and Zhang, Chuanyi and Wang, Yijun and Chen, Ziyun and Xu, Wei and Di, Shimin and Zheng, Yuhui},
  journal={Proceedings of the 33th ACM International conference on Multimedia},
  pages={1-13},
  year={2025}
}

@article{yang2022detecting,
  title={Detecting rotated objects as gaussian distributions and its 3-d generalization},
  author={Yang, Xue and Zhang, Gefan and Yang, Xiaojiang and Zhou, Yue and Wang, Wentao and Tang, Jin and He, Tao and Yan, Junchi},
  journal={IEEE Transactions on Pattern Analysis and Machine Intelligence},
  volume={45},
  number={4},
  pages={4335--4354},
  year={2022},
  publisher={IEEE}
}

@article{anonymous2024textseg,
  title={Text4Seg: Reimagining Image Segmentation as Text Generation},
  author={Lan, Mengcheng and Chen, Chaofeng and Zhou, Yue and Xu, Jiaxing and Ke, Yiping and Wang, Xinjiang and Feng, Litong and Zhang, Wayne},
  journal={International Conference on Learning Representations},
  pages={1-15},
  year={2024}
}

@article{qiao2020referring,
  title={Referring expression comprehension: A survey of methods and datasets},
  author={Qiao, Yanyuan and Deng, Chaorui and Wu, Qi},
  journal={IEEE Transactions on Multimedia},
  volume={23},
  pages={4426--4440},
  year={2020},
  publisher={IEEE}
}

@inproceedings{lai2024lisa,
  title={Lisa: Reasoning segmentation via large language model},
  author={Lai, Xin and Tian, Zhuotao and Chen, Yukang and Li, Yanwei and Yuan, Yuhui and Liu, Shu and Jia, Jiaya},
  booktitle={Proceedings of the IEEE/CVF Conference on Computer Vision and Pattern Recognition},
  pages={9579--9589},
  year={2024}
}

@article{li2025segearth,
  title={Segearth-r1: Geospatial pixel reasoning via large language model},
  author={Li, Kaiyu and Xin, Zepeng and Pang, Li and Pang, Chao and Deng, Yupeng and Yao, Jing and Xia, Guisong and Meng, Deyu and Wang, Zhi and Cao, Xiangyong},
  journal={arXiv preprint arXiv:2504.09644},
  year={2025}
}

@article{gu2021proximal,
  title={Proximal policy optimization with policy feedback},
  author={Gu, Yang and Cheng, Yuhu and Chen, CL Philip and Wang, Xuesong},
  journal={IEEE Transactions on Systems, Man, and Cybernetics: Systems},
  volume={52},
  number={7},
  pages={4600--4610},
  year={2021},
  publisher={IEEE}
}

@article{liu2025seg,
  title={Seg-zero: Reasoning-chain guided segmentation via cognitive reinforcement},
  author={Liu, Yuqi and Peng, Bohao and Zhong, Zhisheng and Yue, Zihao and Lu, Fanbin and Yu, Bei and Jia, Jiaya},
  journal={arXiv preprint arXiv:2503.06520},
  year={2025}
}

@InProceedings{muhtar2024lhrs,
author="Muhtar, Dilxat and Li, Zhenshi and Gu, Feng and Zhang, Xueliang and Xiao, Pengfeng",
title="LHRS-Bot: Empowering Remote Sensing with VGI-Enhanced Large Multimodal Language Model",
booktitle="Computer Vision -- ECCV 2024",
year="2025",
publisher="Springer Nature Switzerland",
address="Cham",
pages="440--457",
isbn="978-3-031-72904-1"
}

@article{wu2024deepseek,
	author = {Guo, Daya and Yang, Dejian and Zhang, Haowei and Song, Junxiao and Wang, Peiyi and Zhu, Qihao, et al.},
	date = {2025/09/01},
	id = {Guo2025},
	isbn = {1476-4687},
	journal = {Nature},
	number = {8081},
	pages = {633--638},
	title = {DeepSeek-R1 incentivizes reasoning in LLMs through reinforcement learning},
	volume = {645},
	year = {2025}}

@inproceedings{chen2024internvl,
  title={Internvl: Scaling up vision foundation models and aligning for generic visual-linguistic tasks},
  author={Chen, Zhe and Wu, Jiannan and Wang, Wenhai and Su, Weijie and Chen, Guo and Xing, Sen and Zhong, Muyan and Zhang, Qinglong and Zhu, Xizhou and Lu, Lewei and others},
  booktitle={Proceedings of the IEEE/CVF conference on computer vision and pattern recognition},
  pages={24185--24198},
  year={2024}
}

@article{bai2025qwen2,
  title={Qwen2. 5-vl technical report},
  author={Bai, Shuai and Chen, Keqin and Liu, Xuejing and Wang, Jialin and Ge, Wenbin and Song, Sibo and Dang, Kai and Wang, Peng and Wang, Shijie and Tang, Jun and others},
  journal={arXiv preprint arXiv:2502.13923},
  year={2025}
}

@inproceedings{ren2024pixellm,
  title={Pixellm: Pixel reasoning with large multimodal model},
  author={Ren, Zhongwei and Huang, Zhicheng and Wei, Yunchao and Zhao, Yao and Fu, Dongmei and Feng, Jiashi and Jin, Xiaojie},
  booktitle={Proceedings of the IEEE/CVF Conference on Computer Vision and Pattern Recognition},
  pages={26374--26383},
  year={2024}
}

@InProceedings{zhang2023next,
  title = 	 {{NE}x{T}-Chat: An {LMM} for Chat, Detection and Segmentation},
  author =       {Zhang, Ao and Yao, Yuan and Ji, Wei and Liu, Zhiyuan and Chua, Tat-Seng},
  journal = 	 {Proceedings of the 41st International Conference on Machine Learning},
  pages = 	 {60116--60133},
  year = 	 {2024},
  volume = 	 {235},
  series = 	 {Proceedings of Machine Learning Research},
  month = 	 {21--27 Jul},
}

@ARTICLE{ou2025geopix,
  author={Ou, Ruizhe and Hu, Yuan and Zhang, Fan and Chen, Jiaxin and Liu, Yu},
  journal={IEEE Geoscience and Remote Sensing Magazine}, 
  title={GeoPix: A multimodal large language model for pixel-level image understanding in remote sensing}, 
  year={2025},
  volume={13},
  number={3},
  pages={324-337}}

@article{shabbir2025geopixel,
  title={Geopixel: Pixel grounding large multimodal model in remote sensing},
  author={Shabbir, Akashah and Zumri, Mohammed and Bennamoun, Mohammed and Khan, Fahad S and Khan, Salman},
  journal={International Conference on Machine Learning},
  pages={1-13},
  year={2025}
}

@inproceedings{pang2025vhm,
  title={Vhm: Versatile and honest vision language model for remote sensing image analysis},
  author={Pang, Chao and Weng, Xingxing and Wu, Jiang and Li, Jiayu and Liu, Yi and Sun, Jiaxing and Li, Weijia and Wang, Shuai and Feng, Litong and Xia, Gui-Song and others},
  booktitle={Proceedings of the AAAI Conference on Artificial Intelligence},
  volume={39},
  number={6},
  pages={6381--6388},
  year={2025}
}

@inproceedings{wu2020phrasecut,
  title={Phrasecut: Language-based image segmentation in the wild},
  author={Wu, Chenyun and Lin, Zhe and Cohen, Scott and Bui, Trung and Maji, Subhransu},
  booktitle={Proceedings of the IEEE/CVF Conference on Computer Vision and Pattern Recognition},
  pages={10216--10225},
  year={2020}
}

@misc{zhao2024swift,
      title={SWIFT:A Scalable lightWeight Infrastructure for Fine-Tuning},
      author={Yuze Zhao and Jintao Huang and Jinghan Hu and Xingjun Wang and Yunlin Mao and Daoze Zhang and Zeyinzi Jiang and Zhikai Wu and Baole Ai and Ang Wang and Wenmeng Zhou and Yingda Chen},
      year={2024},
      eprint={2408.05517},
      archivePrefix={arXiv},
      primaryClass={cs.CL}
}

@inproceedings{kirillov2023segment,
  title={Segment anything},
  author={Kirillov, Alexander and Mintun, Eric and Ravi, Nikhila and Mao, Hanzi and Rolland, Chloe and Gustafson, Laura and Xiao, Tete and Whitehead, Spencer and Berg, Alexander C and Lo, Wan-Yen and others},
  booktitle={Proceedings of the IEEE/CVF international conference on computer vision},
  pages={4015--4026},
  year={2023}
}

@article{mou2020era,
title={ERA: A data set and deep learning benchmark for event recognition in aerial videos [software and data sets]},
author={Mou, Lichao and Hua, Yuansheng and Jin, Pu and Zhu, Xiao Xiang},
journal={IEEE Geoscience and Remote Sensing Magazine},
volume={8},
number={4},
pages={125--133},
year={2020},
publisher={IEEE}
}

@article{zhu2021detection,
  title={Detection and tracking meet drones challenge},
  author={Zhu, Pengfei and Wen, Longyin and Du, Dawei and Bian, Xiao and Fan, Heng and Hu, Qinghua and Ling, Haibin},
  journal={IEEE Transactions on Pattern Analysis and Machine Intelligence},
  volume={44},
  number={11},
  pages={7380--7399},
  year={2021},
  publisher={IEEE}
}

@article{hollich2007young,
  title={Young children associate novel words with complex objects rather than salient parts.},
  author={Hollich, George and Golinkoff, Roberta M and Hirsh-Pasek, Kathy},
  journal={Developmental Psychology},
  volume={43},
  number={5},
  pages={1051},
  year={2007},
  publisher={American Psychological Association}
}

@article{smallman1996foundations,
  title={Foundations of Vision, Cognitive Psychology},
  author={Smallman, H. S.},
  journal={Perception},
  volume={25},
  number={6},
  pages={751--754},
  year={1996}
}

@article{osina2017out,
  title={Out of reach, out of mind? Infants’ comprehension of references to hidden inaccessible objects},
  author={Osina, Maria A and Saylor, Megan M and Ganea, Patricia A},
  journal={Child development},
  volume={88},
  number={5},
  pages={1572--1580},
  year={2017},
  publisher={Wiley Online Library}
}

@article{zhou2025airspatial,
  author={Zhou, Yue and Ding, Ran and Yang, Xue and Jiang, Xue and Liu, Xingzhao},
  journal={IEEE Transactions on Geoscience and Remote Sensing}, 
  title={AirSpatialBot: A Spatially Aware Aerial Agent for Fine-Grained Vehicle Attribute Recognition and Retrieval}, 
  year={2025},
  volume={63},
  number={},
  pages={1-12},
  }

@inproceedings{wang2025xlrs,
  title={Xlrs-bench: Could your multimodal llms understand extremely large ultra-high-resolution remote sensing imagery?},
  author={Wang, Fengxiang and Wang, Hongzhen and Guo, Zonghao and Wang, Di and Wang, Yulin and Chen, Mingshuo and Ma, Qiang and Lan, Long and Yang, Wenjing and Zhang, Jing and others},
  booktitle={Proceedings of the Computer Vision and Pattern Recognition Conference},
  pages={14325--14336},
  year={2025}
}

@article{yang2025look,
  title={Look-back: Implicit visual re-focusing in mllm reasoning},
  author={Yang, Shuo and Niu, Yuwei and Liu, Yuyang and Ye, Yang and Lin, Bin and Yuan, Li},
  journal={arXiv preprint arXiv:2507.03019},
  year={2025}
}

@article{jaccard1901etude,
  title={{\'E}tude comparative de la distribution florale dans une portion des Alpes et des Jura},
  author={Jaccard, Paul},
  journal={Bull Soc Vaudoise Sci Nat},
  volume={37},
  pages={547--579},
  year={1901}
}

@article{peng2023kosmos,
  title={Kosmos-2: Grounding multimodal large language models to the world},
  author={Peng, Zhiliang and Wang, Wenhui and Dong, Li and Hao, Yaru and Huang, Shaohan and Ma, Shuming and Wei, Furu},
  journal={arXiv preprint arXiv:2306.14824},
  year={2023}
}

@article{chen2023shikra,
  title={Shikra: Unleashing multimodal llm's referential dialogue magic},
  author={Chen, Keqin and Zhang, Zhao and Zeng, Weili and Zhang, Richong and Zhu, Feng and Zhao, Rui},
  journal={arXiv preprint arXiv:2306.15195},
  year={2023}
}

@article{deng2023transvg,
  author={Deng, Jiajun and Yang, Zhengyuan and Liu, Daqing and Chen, Tianlang and Zhou, Wengang and Zhang, Yanyong and Li, Houqiang and Ouyang, Wanli},
  journal={IEEE Transactions on Pattern Analysis and Machine Intelligence}, 
  title={TransVG++: End-to-End Visual Grounding With Language Conditioned Vision Transformer}, 
  year={2023},
  volume={45},
  number={11},
  pages={13636-13652}}

@article{shi2024mdetr,
  author={Shi, Fengyuan and Gao, Ruopeng and Huang, Weilin and Wang, Limin},
  journal={IEEE Transactions on Pattern Analysis and Machine Intelligence}, 
  title={Dynamic MDETR: A Dynamic Multimodal Transformer Decoder for Visual Grounding}, 
  year={2024},
  volume={46},
  number={2},
  pages={1181-1198}}

@InProceedings{refcoco,
author="Yu, Licheng
and Poirson, Patrick
and Yang, Shan
and Berg, Alexander C.
and Berg, Tamara L.",
editor="Leibe, Bastian
and Matas, Jiri
and Sebe, Nicu
and Welling, Max",
title="Modeling Context in Referring Expressions",
booktitle="Computer Vision -- ECCV 2016",
year="2016",
publisher="Springer International Publishing",
address="Cham",
pages="69--85",
}

@InProceedings{gres,
  author={Liu, Chang and Ding, Henghui and Jiang, Xudong},
  booktitle={2023 IEEE/CVF Conference on Computer Vision and Pattern Recognition (CVPR)}, 
  title={GRES: Generalized Referring Expression Segmentation}, 
  year={2023},
  volume={},
  number={},
  pages={23592-23601}}

@InProceedings{li2023irvg,
  author={Li, Menghao and Wang, Chunlei and Feng, Wenquan and Lyu, Shuchang and Cheng, Guangliang and Li, Xiangtai and Liu, Binghao and Zhao, Qi},
  booktitle={2023 IEEE/CVF International Conference on Computer Vision Workshops (ICCVW)}, 
  title={Iterative Robust Visual Grounding with Masked Reference based Centerpoint Supervision}, 
  year={2023},
  volume={},
  number={},
  pages={4653-4658}}

@ARTICLE{wu2024rris,
  author={Wu, Jianzong and Li, Xiangtai and Li, Xia and Ding, Henghui and Tong, Yunhai and Tao, Dacheng},
  journal={IEEE Transactions on Image Processing}, 
  title={Toward Robust Referring Image Segmentation}, 
  year={2024},
  volume={33},
  number={},
  pages={1782-1794}}



\end{document}